\documentclass{article}

\usepackage{microtype}
\usepackage{graphicx}
\usepackage{subcaption}
\usepackage{booktabs} 
\usepackage{multirow}

\usepackage{xurl}
\usepackage{hyperref}

\usepackage[preprint]{icml2026}

\usepackage{amsmath}
\usepackage{amssymb}
\usepackage{mathtools}
\usepackage{amsthm}
\usepackage{enumitem}

\usepackage[capitalize,noabbrev]{cleveref}

\theoremstyle{plain}

\theoremstyle{definition}

\theoremstyle{remark}

\usepackage[textsize=tiny]{todonotes}

\icmltitlerunning{From Atoms to Trees: Building a Structured Feature Forest with Hierarchical Sparse Autoencoders}

\begin{document}

\twocolumn[
  \icmltitle{From Atoms to Trees: Building a Structured Feature Forest\\ with Hierarchical Sparse Autoencoders}



  \icmlsetsymbol{equal}{*}

  \begin{icmlauthorlist}
    \icmlauthor{Yifan Luo}{equal,sms}
    \icmlauthor{Yang Zhan}{equal,sms}
    \icmlauthor{Jiedong Jiang}{westlakeITS}
    \icmlauthor{Tianyang Liu}{uoe}
    \icmlauthor{Mingrui Wu}{bit}
    \icmlauthor{Zhennan Zhou}{sswu}
    \icmlauthor{Bin Dong}{bicmr,cmlr}
  \end{icmlauthorlist}

  \icmlaffiliation{sms}{School of Mathematical Science, Peking University, Beijing, China}
  \icmlaffiliation{bicmr}{Beijing International Center for Mathematical Research and the New Cornerstone
Science Laboratory, Peking University, Beijing, China}
  \icmlaffiliation{cmlr}{Center for Machine Learning Research, Peking University, Beijing, China}
  \icmlaffiliation{bit}{School of Information and Electronics, Beijing Institute of Technology, Beijing, China}
  \icmlaffiliation{westlakeITS}{Institute for Theoretical Sciences, Westlake University, Hangzhou, China}
  \icmlaffiliation{sswu}{School of Science, Westlake University, Hangzhou, China}
  \icmlaffiliation{uoe}{School of Informatics, University of Edinburgh, Edinburgh, UK}

  \icmlcorrespondingauthor{Yifan Luo}{luoyf@pku.edu.cn}

  \icmlkeywords{ICML, Mechanism Interpretability, Sparse Autoencoder}

  \vskip 0.3in
]



\printAffiliationsAndNotice{}  

\begin{abstract}
Sparse autoencoders (SAEs) have proven effective for extracting monosemantic features from large language models (LLMs), yet these features are typically identified in isolation. However, broad evidence suggests that LLMs capture the intrinsic structure of natural language, where the phenomenon of ``feature splitting'' in particular indicates that such structure is hierarchical. To capture this, we propose the \textbf{Hierarchical Sparse Autoencoder (HSAE)}, which jointly learns a series of SAEs and the parent-child relationships between their features. HSAE strengthens the alignment between parent and child features through two novel mechanisms: a structural constraint loss and a random feature perturbation mechanism. Extensive experiments across various LLMs and layers demonstrate that HSAE consistently recovers semantically meaningful hierarchies, supported by both qualitative case studies and rigorous quantitative metrics. At the same time, HSAE preserves the reconstruction fidelity and interpretability of standard SAEs across different dictionary sizes. Our work provides a powerful, scalable tool for discovering and analyzing the multi-scale conceptual structures embedded in LLM representations.
\end{abstract}

\section{Introduction}

Understanding the mechanisms by which large language models (LLMs) represent and process information is of great interest in the machine learning community. Recent advances in mechanistic interpretability have shifted focus from analyzing individual neurons \citep{olah2020zoom,bills2023language}, which are often polysemantic, toward identifying more coherent units of analysis. Sparse autoencoders (SAEs) have emerged as a particularly successful approach in this direction \citep{bricken2023monosemanticity,cunningham2023sparse}. By learning an overcomplete dictionary that maps dense activations to a sparse latent space, SAEs scalably extract features that are human-interpretable.

Beyond identifying isolated features, emerging evidence suggests that SAE representations are not merely atomic collections but are embedded within a sophisticated internal organization. Recent work has revealed spatial clustering corresponding to semantic groups \citep{li2025geometry} and co-activation patterns between abstract features and their specialized counterparts \citep{clarke2024compositionality}, indicating that LLMs capture linguistic relationships through structured internal representations. Since navigating flat, unstructured SAE dictionaries remains inherently difficult, extracting these latent structures is essential for creating organized, multi-scale interpretations. A particularly notable phenomenon hinting at this organization is \emph{feature splitting}, where broad concepts decompose into more granular sub-features as dictionary size increases \citep{bricken2023monosemanticity}. While prior works have often viewed this as an obstacle to identifying universal, canonical features \citep{leask2025NonCanonical, bussmann2025Matryoshka}, we instead recognize it as direct evidence of the underlying hierarchy represented in the model. By exploiting feature splitting rather than suppressing it, we transform this behavior into a hierarchical indexing structure for the feature space, organizing the ``atoms'' of representation into a structured forest.


\begin{figure*}[ht]
    \centering
    \includegraphics[width=0.85\textwidth]{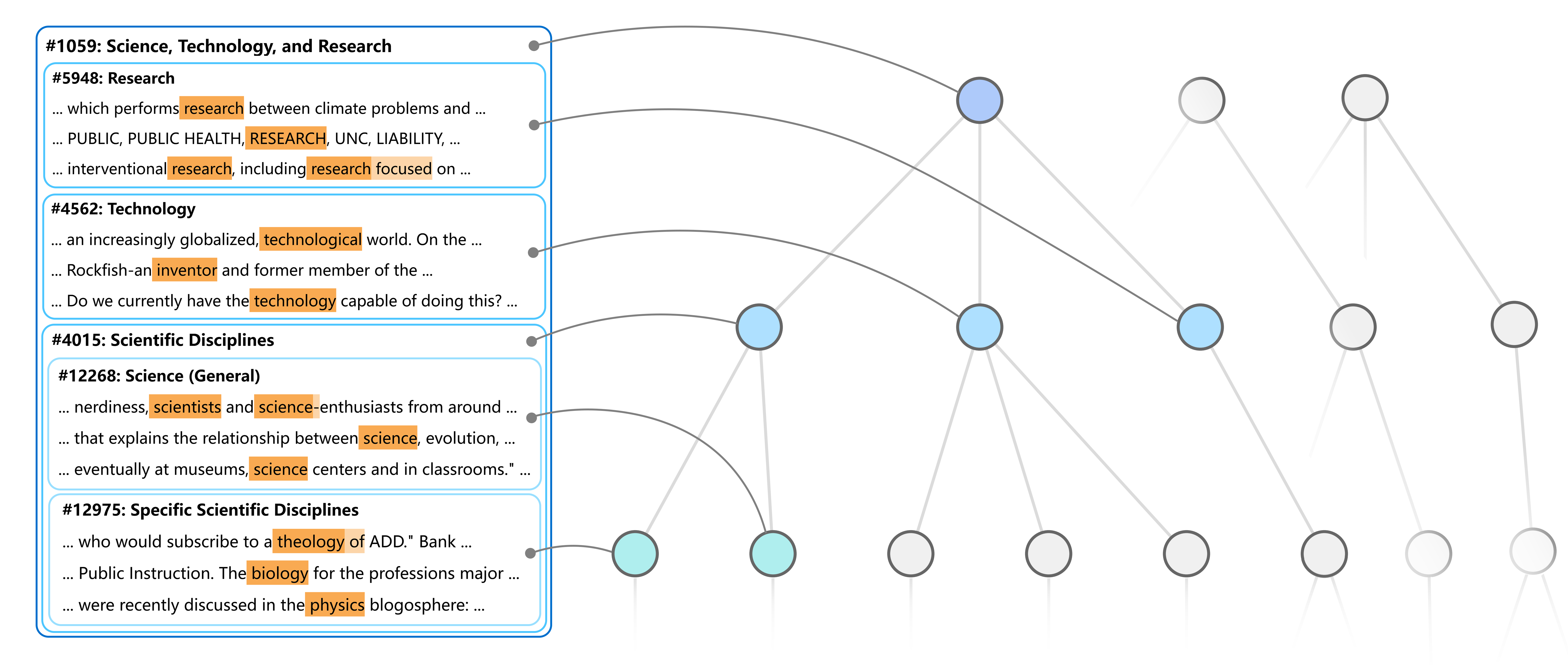}
    \caption{
        \textbf{Hierarchical feature discovery with HSAE.} We visualize a tree within the learned feature forest alongside a semantic dashboard of its nodes. Each node includes a unique feature index (e.g., \#1059), a human-annotated semantic label, and representative top-activating context snippets. Orange highlights indicate the activation positions, with color intensity reflecting the activation magnitude. HSAE captures a clear conceptual taxonomy: a broad parent feature representing \textit{Science, Technology, and Research} (\#1059) is systematically decomposed into specialized features such as \textit{Scientific Disciplines} (\#4015). This intermediate node further branches into a feature capturing lexical patterns containing the word \textit{science} or \textit{scientists} (\#12268) alongside a feature representing specific discipline names such as \textit{biology} or \textit{physics} (\#12975). For clarity, only a three-layer subgraph and examples of some features are shown.
    }
    \label{fig:teaser}
\end{figure*}

For this purpose, we introduce the \textbf{Hierarchical Sparse Autoencoder (HSAE)}. The HSAE simultaneously trains a series of SAE levels with increasing dictionary sizes, with features at each level explicitly assigned to a parent in the preceding coarser level, forming a tree-structured hierarchy. Model parameters and the parent–child assignments co-evolve through alternating optimization. To ensure that hierarchical links reflect semantic subsumption rather than mere structural adjacency, we design optimization objective to enforce functional dependence across levels via: (1) a \textbf{structural constraint loss} encouraging parent features to be reconstructed from their assigned children’s outputs, and (2) a \textbf{random feature perturbation} mechanism that treats parent and children contributions as interchangeable during reconstruction of original activations. These mechanisms align the learned features into a consistent conceptual taxonomy, where higher-level features are incentivized to act as distilled abstractions of their finer-grained descendants.


In our experiments, we apply HSAE to the residual stream activations of different LLMs and find that it captures a rich, multi-layered hierarchical structure among interpretable features. As illustrated in \Cref{fig:teaser}, HSAE naturally organizes discovered features into a conceptual forest, revealing a progressive refinement of semantic meaning. In this structure, high-level parent features initially respond to broad, domain-general patterns, which are then systematically decomposed into increasingly specific features at deeper levels.


To provide a comprehensive evaluation of our results, we first present qualitative case studies, demonstrating that HSAE consistently recovers clear semantic hierarchies across diverse domains. Second, systematic quantitative evaluations confirm that these learned hierarchies are statistically representative of the entire feature space rather than isolated occurrences. Using multiple statistical metrics and an LLM-based automated interpretability assessment, we show that the recovered hierarchical links are both statistically robust and semantically meaningful, significantly outperforming baselines that infer relations post-hoc from independently trained SAEs. Third, we demonstrate that HSAE preserves core SAE performance, matching or exceeding conventional SAEs on standard benchmarks for feature quality and reconstruction fidelity across all dictionary sizes. Finally, we provide extensive ablation studies to validate the necessity of each proposed mechanism.

Beyond fundamental validation, our analysis reveals several intriguing structural properties of the learned feature space. Visualization shows that input activations triggering sibling features are significantly more related in their low-dimensional projections than those triggering unrelated features, indicating that HSAE captures the underlying geometry of conceptual manifolds. Additionally, we find that features with multiple children often exhibit slightly higher interpretability than those with a single child, providing preliminary evidence that the branching factor may serve as a hint for assessing a feature’s semantic clarity. 


The main contributions of this work are as follows:
\begin{itemize}[leftmargin=*]
    \item \textbf{A Novel Hierarchical Architecture:} We propose HSAE, the first SAE framework to integrate structural priors directly into training via structural constraint loss and random feature perturbation, enabling the learning of an organized ``conceptual forest.''
    \item \textbf{Benchmarking Hierarchical Consistency:} We establish a set of benchmarks to measure hierarchy consistency, introducing statistical metrics—such as parent-child co-activation probability—alongside an LLM-based automated interpretability framework. Our results demonstrate that HSAE recovers more robust and coherent structures than post-hoc analysis of independent SAEs.
    \item \textbf{New Insights into LLM Representation Structures:} We provide empirical evidence of how LLMs organize complex concepts into fine-grained hierarchies. Our analysis reveals that these structures possess intrinsic geometric properties, characterized by the spatial clustering of sibling features, and that the structural branching factor serves as a predictor of feature semantic clarity.
\end{itemize}

\section{Related Works}

\subsection{Discovering Structures in Model Activations}

Early analyses of word embedding models identified the well-known parallelogram rule, where semantic analogies are captured through linear vector arithmetic such as $v(\texttt{"king"}) - v(\texttt{"man"}) + v(\texttt{"woman"}) \approx v(\texttt{"queen"})$ \citep{mikolov2013distributed,mikolov2013linguistic}. Similar phenomena have been observed in the activations of LLMs \citep{gurnee2024language,heinzerlinginui2024monotonic}. These observations support the linear representation hypothesis, which posits that the representation of semantic concepts (features) in the activation space of LLMs corresponds to one-dimensional directions \citep{park2023LRH}. This hypothesis motivated the development of SAEs, which decompose a model's activations into an overcomplete basis of interpretable linear features \citep{bricken2023monosemanticity, cunningham2023sparse}.

SAE research has revealed various forms of spatial structure within the feature space. Initial work identified \emph{feature splitting}, where broad concepts break into more granular sub-features \cite{bricken2023monosemanticity}. Further geometric analyses found multi-scale cluster structures in the latent space corresponding to functional domains and semantic groups such as mathematics, code, and natural language \citep{li2025geometry}. Research has also uncovered irreducible two-dimensional circular manifolds that SAEs use to represent periodic concepts like weekdays \citep{engels2025not}. Moving beyond static spatial organization, recent work explores feature structures through co-activation patterns, revealing hierarchical ``hub-and-spoke'' topologies in co-occurrence graphs, where abstract features act as central hubs connected to more specialized spokes \citep{clarke2024compositionality}.


\subsection{Feature Splitting}

Feature splitting \citep{bricken2023monosemanticity} is an observed phenomenon in SAEs where an interpretable, broad feature decomposes into multiple specialized ones as the dictionary size increases. For example, a general ``punctuation mark'' feature may split into distinct latents for periods, commas, and question marks. Related phenomena that can affect feature quality include \emph{feature composition}, where frequently co-occurring independent features merge into a single latent \citep{anders2024composedtoymodels,wattenberg2024relational}, and \emph{feature absorption}, where a general feature develops blind spots for subcases captured by more specialized latents \citep{Chanin2024AIF}. These effects can be explained by the sparsity-driven training objectives of SAEs \citep{Chanin2024AIF}. Metrics have been proposed to quantitatively measure these effects within an SAE \citep{Huang2024RAVELEI,Chanin2024AIF,Karvonen2025SAEBenchAC}.


Several recent works treat feature splitting phenomena as drawbacks \citep{bussmann2025Matryoshka, chanin2025feature, korznikov2025ortsae}. For instance, Matryoshka SAEs address these issues by simultaneously training nested dictionaries of increasing size, forcing smaller dictionaries to reconstruct inputs independently. This approach is designed to preserve high-level concepts in smaller dictionaries while allowing larger dictionaries to learn specializations. In contrast to viewing splitting solely as a defect, our work interprets it as evidence of an underlying hierarchical conceptual structure in the model and aims to extract this hierarchy.

\subsection{Architectural Evolution of Sparse Autoencoders}

The engineering of SAEs has evolved from resolving practical training bottlenecks toward architectural designs that address high-level feature quality issues. Early SAEs used \( L_1 \) regularization with ReLU activations \citep{bricken2023monosemanticity,cunningham2023sparse}, but suffered from feature shrinkage.
Subsequent variants addressed this issue: Gated SAEs \citep{rajamanoharan2024gate} decoupled detection from magnitude estimation, while TopK \citep{gao2025scaling} and BatchTopK \citep{bussmann2024batchtopk} enforced hard \( L_0 \) sparsity constraints via ghost gradients. Training dynamics and reconstruction fidelity were improved through JumpReLU SAEs \citep{Rajamanoharan2024JumpingAI}, which introduced learnable activation thresholds, and \( p \)-annealing techniques \citep{Karvonen2024Measuring} that interpolate between \( L_1 \) and \( L_0 \) objectives. For large-scale efficiency, Switch SAEs \citep{mudide2025efficient} adopted mixture-of-experts routing. Recent structural innovations include Meta-SAEs \citep{leask2025NonCanonical}, which decompose the composite features of another SAE into more fundamental units, and Matryoshka SAEs \citep{bussmann2025Matryoshka}, which address feature absorption and splitting through nested dictionaries. Subsequent refinements aim to further reduce feature redundancy through distillation cycles \citep{martin2025attribution} and to reduce correlated feature merging in smaller SAEs \citep{chanin2025feature}. However, approaches that explicitly learn structures within the feature space remain underexplored.


\section{Hierarchical Sparse Autoencoders}

In HSAE, we jointly trains a collection of SAEs of increasing dictionary size and explicitly enforces a hierarchical structure among their features.

Technically, a standard SAE seeks to decompose the activation vector $x \in \mathbb{R}^d$ into a sparse linear combination of interpretable directions. Let $\{\operatorname{SAE}_\ell (\cdot)\}_{\ell=1}^L$ denote a sequence of $L$ SAEs, where
{
\small
$$
\operatorname{SAE}_\ell (x)=\sum_{i=1}^{n_\ell} d_{\ell,i}\,\sigma(e_{\ell,i}^T x).
$$
}
Later in the paper, we also use $f_{\ell,i}(x)=d_{\ell,i}\,\sigma(e_{\ell,i}^T x)$ to denote a single feature function. Here, $\ell$ indexes the hierarchy level and $n_\ell$ represent the dictionary size of level $\ell$. Each $\operatorname{SAE}_\ell$ is a standard two-layer JumpRELU SAE \cite{Rajamanoharan2024JumpingAI}. For simplicity, we omit the threshold parameter in $\sigma$, which is also different in each feature $(\ell, i)$. Note that the hierarchy index $\ell$ does \emph{not} correspond to depth in a multilayer neural networks.

HSAE induces a partial tree structure over features by defining parent--child relationships between selected features in adjacent levels. For each feature $(\ell,i)$ where $1\leqslant \ell\leqslant L-1$, we define a set of children indices
$$
\mathcal{C}_{(\ell,i)} \subseteq \{1,\ldots,n_{\ell+1}\}.
$$
We refer to $(\ell,i)$ as a parent feature and $\{(\ell+1,j)\mid j\in\mathcal{C}_{(\ell,i)}\}$ as its children. To ensure a tree-structured hierarchy, we require that child sets are disjoint across parent features
$$
\mathcal{C}_{(\ell,i)} \cap \mathcal{C}_{(\ell,k)} = \varnothing,\quad \forall\ i \neq k.
$$
The hierarchical relations introduced above do not by themselves impose constraints among related features.
Instead, they define a structural prior that is instantiated through additional loss terms and auxiliary training mechanisms described later in this section.

\subsection{Loss Function}

Each level’s SAE employs a standard loss, which consists of a MSE reconstruction loss and an $L_0$ sparsity penalty
{\small
\begin{equation*}
\mathcal{L}_{\text{SAE},\ell}(x) = \Vert \operatorname{SAE}_\ell(x)-x \Vert_2^2+\lambda_\ell \sum_{i=1}^{n_i} 1_{\{\sigma(e_{\ell,i}^T x)>0\}}.
\end{equation*}
}

For each parent-children group, we introduce a parent-children constraint loss as
{
\small
$$
\mathcal{L}_{\text{PC},(\ell,i)} (x) =
\begin{cases}
    \Vert f_{\ell,i}(x)-\sum_{j\in C_{\ell,i}} f_{\ell+1,j}(x) \Vert_2^2,\ & C_{\ell,i}\neq\varnothing,\\
    0,\ & C_{\ell,i}=\varnothing.
\end{cases}
$$
}
The final HSAE loss function is
{
\small
$$
\mathcal{L}_{\text{HSAE}} = \mathbb{E}_{x\sim D}\left[ \sum_{\ell=1}^L \mathcal{L}_{\text{SAE},\ell}(x) + \rho \left(\sum_{\ell=1}^{L-1} \sum_{i=1}^{n_\ell} \mathcal{L}_{\text{PC},(\ell,i)}(x)\right)\right].
$$
}
where $x\sim D$ is the distribution of activations extracted from the specific site of the target LLM.

\subsection{Parent-Children Feature Perturbations}


Beyond additional loss terms, we also introduce a structure-aware random perturbation of features to encourage the relationship between parent feature and its children.

When computing the SAE loss $\mathcal{L}_{\text{SAE},\ell}$, we do not always use features solely from level $\ell$.
Instead, for each feature, we stochastically substitute its contribution with those of its children at the next level.

Formally, given the perturbation rate $r$, let $z_{\ell,i} \sim \mathrm{Bernoulli}(r)$. We define
{\small
$$
\tilde{f}_{\ell,i}(x)
= z_{\ell,i} f_{\ell,i}(x)
+ (1 - z_{\ell,i}) \sum_{j \in \mathcal{C}_{\ell,i}} f_{\ell+1,j}(x).
$$
}
and compute the SAE loss $\mathcal{L}_{\text{SAE},\ell}$ with the perturbed version of the original SAE:
{
\small
$$
\widetilde{\operatorname{SAE}}_\ell(x) = \sum_{i=1}^{n_\ell} \tilde{f}_{\ell,i}(x)
$$
}
The random variables $z_{\ell,i}$ are sampled independently across features at each training step. This perturbation encourages consistency between parent and children features by exposing the SAE loss to mixtures of representations across adjacent levels.

In practice, we adopt a parent-child constraint weight of $\rho=0.01$ and a perturbation rate of $r=5\%$. These values are selected based on the ablation results in \Cref{fig:hyperpara} to balance hierarchical alignment with reconstruction fidelity.

\subsection{Optimization Strategy}

Training HSAE involves the joint optimization of model parameters $\{e_{\ell,i}, d_{\ell,i}, \sigma_{\ell,i}\}$ and the set of hierarchical relations $\{\mathcal{C}_{\ell,i}\}$ simultaneously. To handle this joint objective, we adopt an alternating optimization strategy. Specifically, the training process alternates between the following two stages:
\begin{itemize}[leftmargin=*]
\item \textbf{Parameter Optimization}: With the hierarchical structure fixed, we optimize the model parameters via gradient descend for a predefined number of steps.
\item \textbf{Hierarchy Update}: With the model parameters fixed, we update the hierarchical structure $\{\mathcal{C}_{\ell,i}\}$ to reflect the current feature organization.
\end{itemize}
This iterative process is maintained throughout training, ensuring that the hierarchical structure co-evolves with the learned SAE features. In practice, we perform a round of hierarchy update every 5000 gradient steps.

\paragraph{Parameter Optimization.}
Most settings for parameter optimization in HSAE is identical to standard JumpReLU SAE training. We use a straight-through estimator of the gradient through the discontinuity of $L_0$ sparsity penalty.
Decoder vectors are normalized after each gradient step to fix their norms.
We also adopt ghost gradients to mitigate the dead neuron problem~\cite{anthropic2024january}.

The only notable modification we introduce is the adaptive sparsity weight. We dynamically adjust the sparsity weights $\lambda_\ell$ during training to guide the average $L_0$ of each level toward a shared target sparsity. The detailed method is provided in Appendix~\ref{app:exp:setup}.

\paragraph{Similarity-Based Hierarchy Update.}
The hierarchy update step is guided by a general principle of feature similarity.
Each feature at level $\ell+1$ will select a new parent feature at level $\ell$ that has the most similarity according to a predefined similarity metric.

We consider several instantiations of this principle.
Similarity can be defined using feature co-activation statistics, measured as the proportion of training examples on which two features are jointly active, with statistics tracked via an exponential moving average.
Alternatively, similarity can be computed in parameter space using the cosine similarity between encoder or decoder feature vectors.
Empirically, HSAE is largely insensitive to the specific choice of similarity metric, with all variants exhibiting comparable behavior in ablation studies (\Cref{fig:similarity_method}). Unless otherwise specified, we use encoder vector similarity throughout this paper.

\paragraph{Partial Tree Structure.}
\label{sec:exp}
Recognizing that not all features naturally conform to a hierarchical taxonomy, we relax the rigid tree constraint by allowing a subset of child features to remain unassigned. Specifically, we implement a quantile-based filtering mechanism where features whose maximum similarity scores fall within the given bottom percentile of all candidates are excluded from the hierarchy. We treat this exclusion level as a tunable hyperparameter, which is set to 20\% in our experiments. This partial tree structure provides the necessary flexibility to accommodate independent features—those capturing idiosyncratic or atomic concepts—without forcing them into incoherent parent-child relationships, thereby preserving the structural faithfulness of the discovered hierarchy.

\section{Experiments}

In this section, we present a comprehensive evaluation of HSAE, combining qualitative observations with rigorous quantitative analysis to demonstrate its ability to capture a significantly richer internal organization than standard SAEs. We begin by providing qualitative evidence of the interpretable hierarchies learned by HSAE, followed by a series of evaluations using statistical metrics and automated interpretability assessments to confirm that these structures are recovered consistently across the model. To ensure that this hierarchical organization does not compromise core performance, we further benchmark HSAE against established SAE evaluation protocols, measuring its feature quality and reconstruction fidelity relative to standard baselines. Finally, we conduct ablation studies to isolate how explicit constraint and implicit perturbation independently and jointly contribute to the formation of semantically coherent and statistically robust hierarchies.

\paragraph{Experimental Setup.}
Unless otherwise specified, all models discussed in this section are trained on 100M residual stream activations extracted from the 13th layer of \texttt{gemma2-2b}~\cite{Riviere2024Gemma2I} as it processes the mini-PILE dataset~\cite{Kaddour2023TheMC}. For both HSAEs and standard SAE baselines, we train four levels of SAEs with dictionary sizes of 2048, 4096, 8192, and 16384. We target a sparsity of $L_0=50$ for each level. Additional training details are provided in Appendix~\ref{app:exp:setup}.

To ensure the robustness of our findings, we also conduct extensive experiments across different layers and target sparsity. These additional results are provided in Appendix~\ref{app:exp:more_exp}. We observe consistent results across all tested settings.

\subsection{Empirical Case Studies}

To provide intuition on the hierarchical structures captured by HSAE, we first present representative case studies from our learned conceptual forest. As illustrated in Figure~\ref{fig:teaser}, HSAE successfully decomposes an abstract root representing \textit{Science, Technology, and Research} into distinct meso-scale clusters representing its constituent pillars. These clusters further branch into specific features that follow distinct lexical patterns, such as occurrences of the word ``science'' or individual scientific disciplines.

\begin{figure}[ht]
    \centering
    \includegraphics[width=0.80\columnwidth]{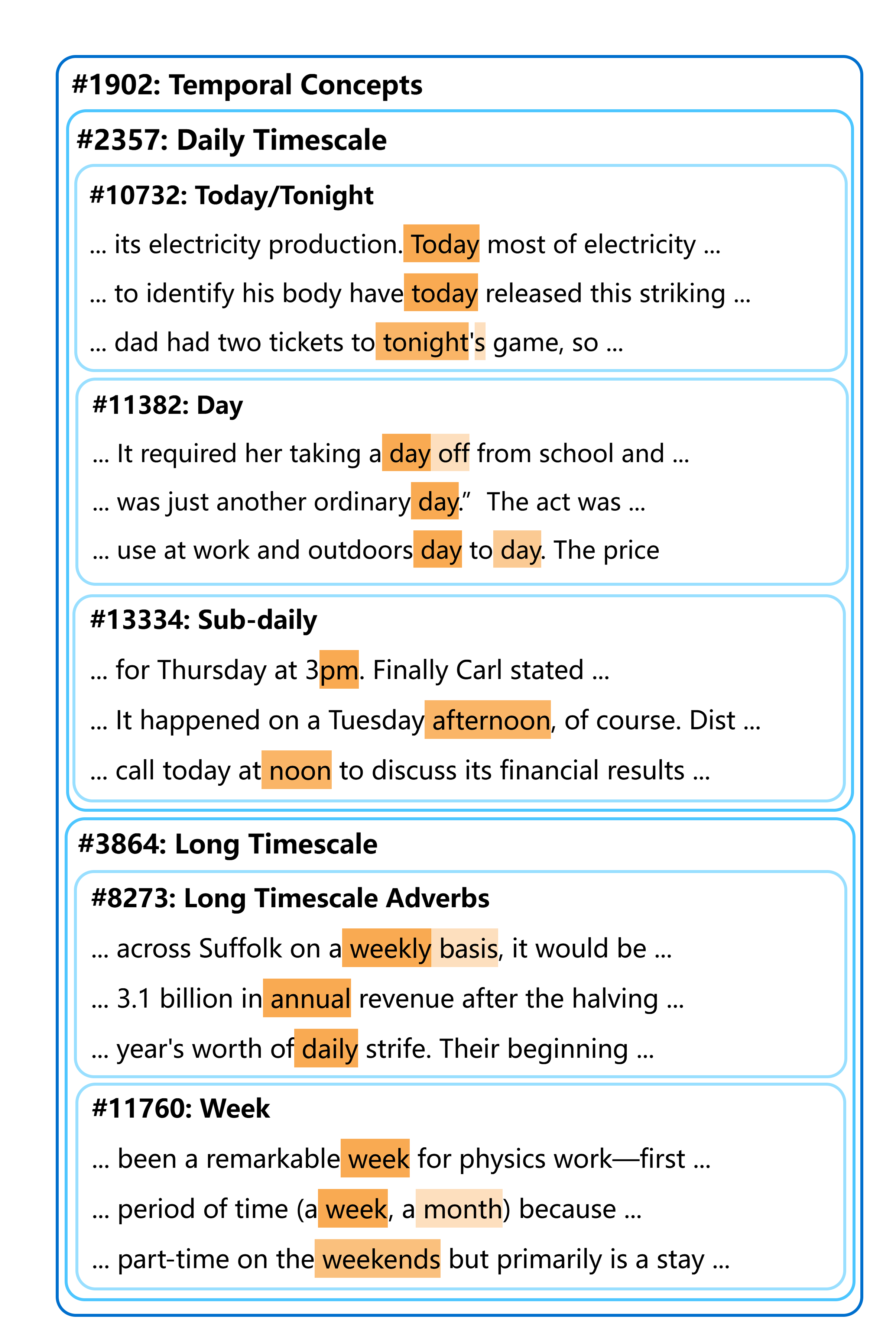}
    \caption{ \textbf{Hierarchical feature discovery of temporal concepts.} The root broad temporal feature (\#1902) branches into a \textit{Daily Timescale} (\#2357) and a \textit{Longer Timescale} (\#3864). At the next level, the daily timescale branch splits into \textit{Today/Tonight} (\#10732), the unit concept \textit{Day} (\#11382), and \textit{Sub-daily} intervals including morning/afternoon or am/pm (\#13334). Similarly, longer-term concepts are partitioned into \textit{Adverbs} (\#8273) and specific units like \textit{Week} (\#11760). Orange highlights indicate activation positions, with intensity reflecting magnitude.}
    \label{fig:empirical}
\end{figure}

The second example, shown in Figure~\ref{fig:empirical}, examines the conceptual organization of \textit{Time}. Here, a coarse-grained root feature captures general temporal references. In the subsequent level, it is partitioned into features representing daily timescales versus longer-term durations (e.g., weeks, months), reflecting a clear organizing principle aligned with human cognition. The daily timescale branch further resolves into specialized sub-domains, segregating immediate deictic markers (e.g., ``today'') from the occurrences of ``day'' and sub-daily intervals. These examples suggest that the decompositions learned by HSAE are not merely stochastic. Rather, they consistently align with human intuition and possess inherent interpretability across semantic scales. We have provided more empirical studies in Appendix~\ref{app:exp:more_empirical}

However, it is worth noting that the labels assigned to these features are over-simplifications of their actual activation patterns, and counter-examples may exist in certain contexts. Furthermore, while the hierarchy captures strong semantic associations, an input that activates a child feature does not guarantee a corresponding trigger of its parent feature. This phenomenon of hierarchical consistency will be rigorously evaluated in the following sections using the parent-child co-activation probability.




\subsection{Statistical Metrics}

To quantitatively evaluate the structural faithfulness of the discovered hierarchy, we employ three metrics that measure the alignment between parent and child features.

\begin{figure*}[ht]
    \centering
    \begin{subfigure}[b]{0.24\textwidth}
        \centering
        \includegraphics[width=\textwidth]{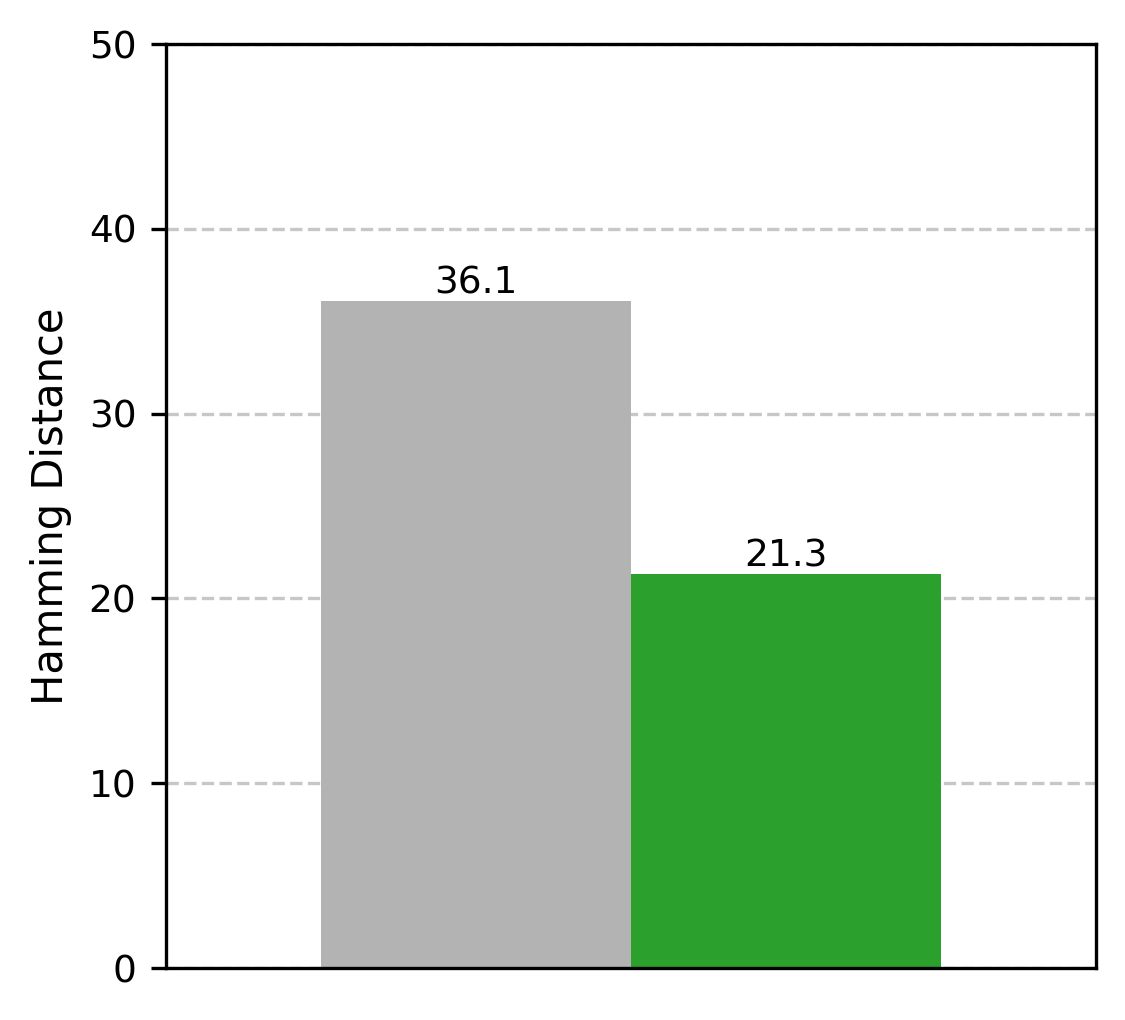}
        \caption{Parent-Child Alignment$\ \downarrow$}
        \label{fig:quant_eval:hamming}
    \end{subfigure}
    \hfill
    \begin{subfigure}[b]{0.24\textwidth}
        \centering
        \includegraphics[width=\textwidth]{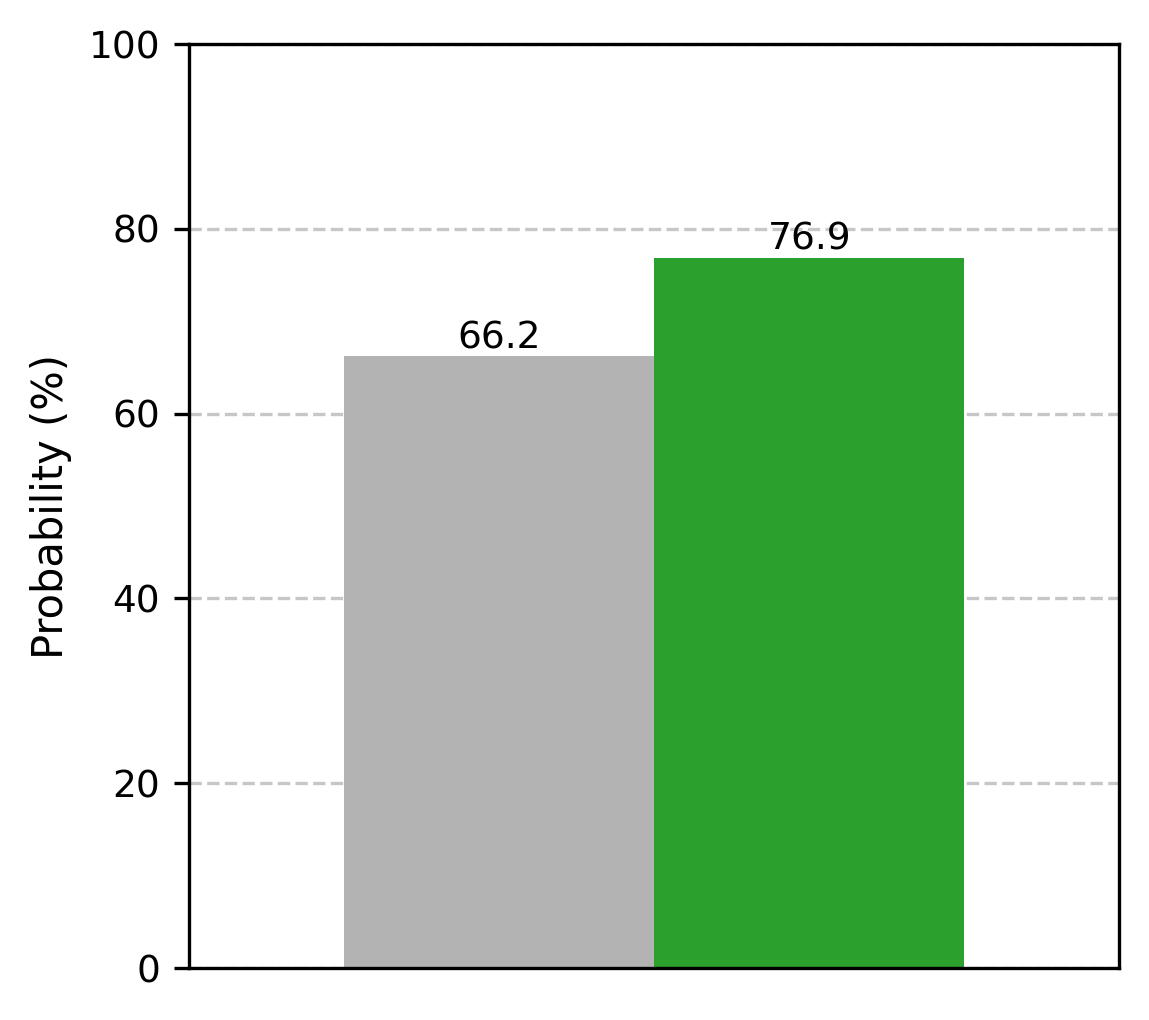}
        \caption{$P(\text{activ}_p \mid \text{activ}_c) \uparrow$}
        \label{fig:quant_eval:child_parent}
    \end{subfigure}
    \hfill
    \begin{subfigure}[b]{0.24\textwidth}
        \centering
        \includegraphics[width=\textwidth]{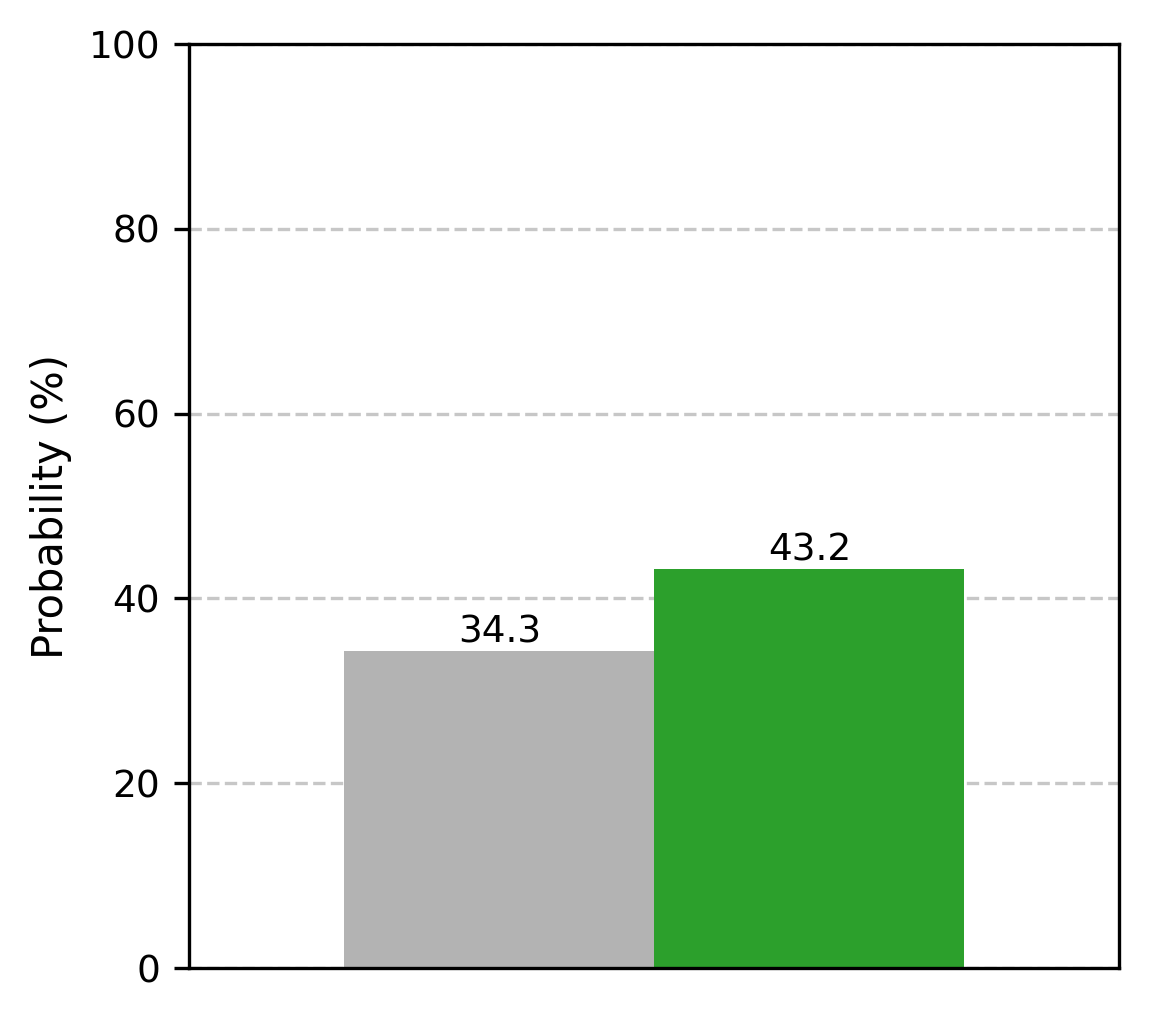}
        \caption{$P(\text{activ}_c \mid \text{activ}_p) \uparrow$}
        \label{fig:quant_eval:parent_child}
    \end{subfigure}
    \hfill
    \begin{subfigure}[b]{0.24\textwidth}
        \centering
        \includegraphics[width=\textwidth]{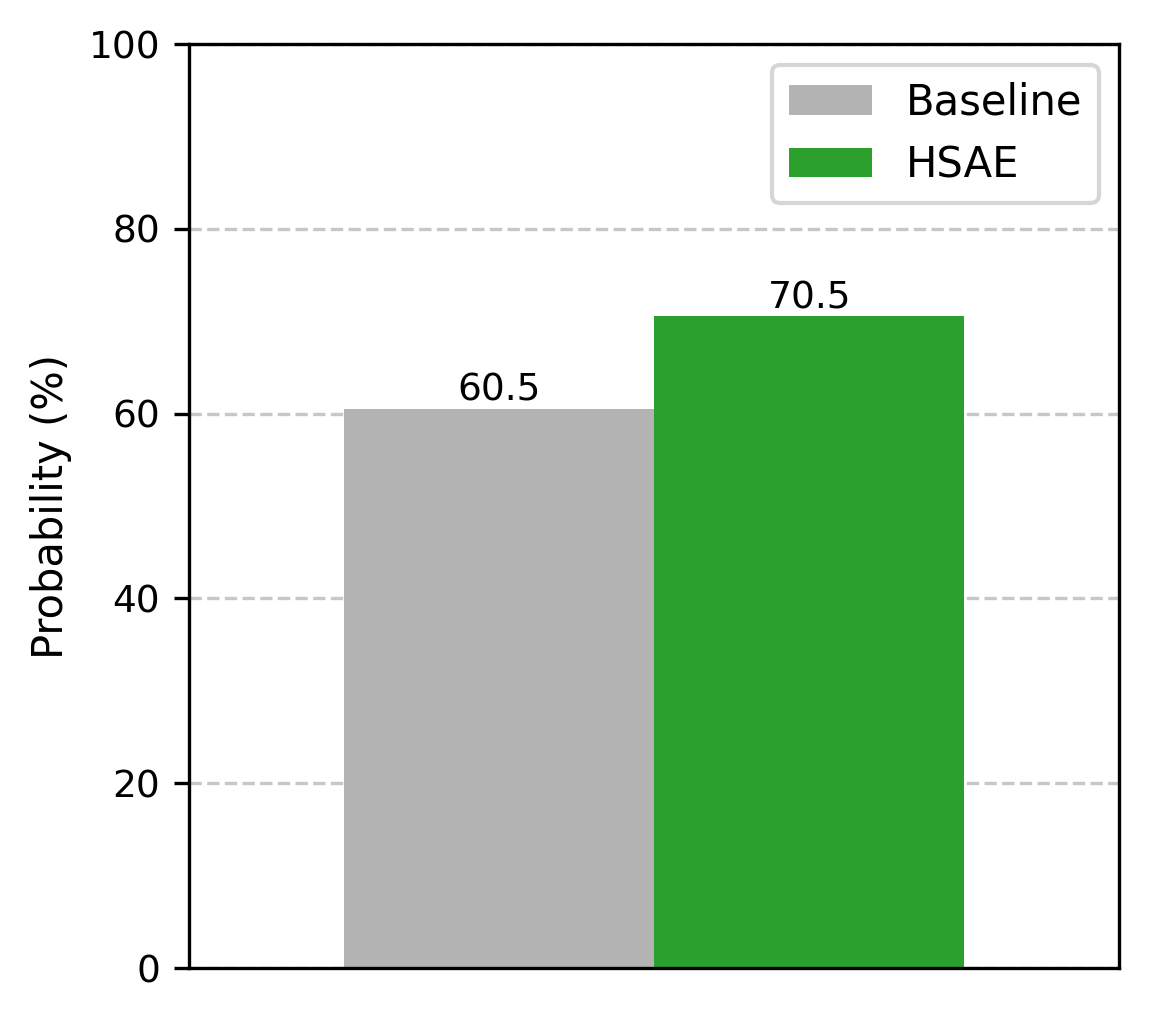}
        \caption{AutoInterp $\uparrow$}
        \label{fig:quant_eval:auto_interp}
    \end{subfigure}
    \caption{
    \textbf{Quantitative evaluation of hierarchical structure.} 
    \textbf{(a) Parent-Child Alignment}: Measured by the Hamming Distance between ground-truth parent activations and their logical-OR reconstruction from children, indicating how well parent features summarize their descendants' activations.
    \textbf{(b) Parent-given-child activation probability}: measuring the necessity of the parent feature for its children. 
    \textbf{(c) Child-given-parent activation probability}: reflecting the coverage of children within the parent's activation space. 
    \textbf{(d) AutoInterp}: LLM-based automated interpretability scores evaluating the semantic alignment between parent and child features.}
    \label{fig:quant_eval}
\end{figure*}

\paragraph{Hierarchical Consistency via Logical-OR.}
We first assess whether parent features serve as effective semantic summaries of their children. We construct a logical-OR prediction where a parent is predicted to be active if at least one of its assigned children is active. We then measure the Hamming distance between this child-based prediction and the ground-truth parent activation pattern (\Cref{fig:quant_eval:hamming}). This metric evaluates the logical alignment of the hierarchy; a lower distance indicates that parent activations are a faithful abstraction of lower-level patterns, confirming that the hierarchy captures a coherent decomposition.


\paragraph{Conditional Co-activation Probabilities.}
Beyond binary logical alignment, we evaluate the statistical dependency between levels through two complementary conditional probabilities, averaged over all assigned parent-child pairs:
\begin{itemize}[leftmargin=*]
\item \textbf{Parent-given-child probability}: Defined as $P(\text{activ}_p \mid \text{activ}_c)=P(\sigma(e_p^Tx)>0 \mid \sigma(e_c^Tx)>0)$, this measures \textbf{hierarchical necessity}. A high probability indicates that a child feature rarely activates without its parent, ensuring that children features remain strictly within the semantic scope of their parents.

\item \textbf{Child-given-parent probability}: Defined as $P(\text{activ}_c \mid \text{activ}_p)=P(\sigma(e_c^Tx)>0 \mid \sigma(e_p^Tx)>0)$, this measures the \textbf{average semantic relevance} of each child feature to its parent. This metric evaluates the statistical strength of individual memberships within the hierarchy, where a higher value suggests that the parent feature is constituted by sub-components that are consistently co-active with it.

\end{itemize}

\paragraph{Baselines.} We compare HSAE against a synthetic feature forest constructed from independently trained SAEs. We train a series of SAEs with dictionary sizes matching the corresponding levels of HSAE, then apply the same similarity-based assignment strategy to build a hierarchy post-hoc. This baseline represents structure obtained via post-hoc feature-splitting approaches.


\paragraph{Results.} As illustrated in \Cref{fig:quant_eval}, HSAE consistently outperforms the post-hoc baseline across all measured metrics. The substantial reduction in Hamming distance indicates that HSAE's parent features provide a significantly clearer and more complete logical summary of their constituent children. Furthermore, the improved co-activation probabilities establish the statistical prerequisite for semantic consistency, ensuring that the discovered hierarchical relations are functionally grounded rather than mere coincidences of vector similarity. These results demonstrate that jointly optimizing hierarchical relations alongside feature parameters enables a more consistent extraction of structure from the activation distribution comparing with post-hoc analysis.

\subsection{Auto-Interpretability of Hierarchies}

To evaluate whether the discovered hierarchical relations correspond to genuine semantic abstractions, we develop an automated pipeline using LLMs as judges. For a given parent-child pair, we independently retrieve high-activation examples for each feature and present them to the LLM. The LLM is tasked with a binary classification: determining whether the two sets of examples exhibit a valid hierarchical conceptual relationship (Yes/No) and providing an associated confidence score. This automated setup allows for a large-scale evaluation of thousands of feature pairs, providing a statistically significant measure of semantic consistency. Detailed sampling strategies, prompt templates, and LLM configurations are provided in Appendix~\ref{app:exp:autointerp}.

\paragraph{Results.} \Cref{fig:quant_eval:auto_interp} reports the rate at which the LLM identifies a valid hierarchical relationship. Again, HSAE significantly outperforms the synthetic baseline, confirming that its joint optimization extracts more semantically grounded structures than post-hoc alignment. This alignment between automated semantic judgment and previous statistical metrics provides strong evidence for the structural faithfulness of HSAE.

\begin{table*}[ht]
    \centering
    \caption{Evluation results for HSAE and baseline SAEs on multiple SAE benchmarks. All metrics except $L_0$ are the higher the better.}
    \begin{tabular}{ccccccccc}
    \toprule
        \textbf{Model} & \textbf{Dict. Size} & \textbf{$L_0$} & \textbf{Var. Exp.} & \textbf{Absorption}& \textbf{AutoInterp} & \textbf{RAVEL} & \textbf{SCR} & \textbf{Sparse Probing} \\
    \midrule
        \multirow{4}{*}{HSAE} 
        & 2048  & 49.4 & 0.612 & 0.983 & 0.807 & 0.521 & 0.272 & 0.868 \\
        & 4096  & 50.0 & 0.651 & 0.981 & 0.814 & 0.582 & 0.286 & 0.873 \\
        & 8192  & 50.5 & 0.685 & 0.951 & 0.859 & 0.627 & 0.314 & 0.874 \\
        & 16384 & 50.6 & 0.710 & 0.922 & 0.869 & 0.654 & 0.324 & 0.873 \\
    \midrule
        \multirow{4}{*}{Baseline}
        & 2048  & 49.4 & 0.613 & 0.988 & 0.813 & 0.539 & 0.245 & 0.863 \\
        & 4096  & 49.8 & 0.651 & 0.974 & 0.829 & 0.609 & 0.302 & 0.879 \\
        & 8192  & 50.3 & 0.684 & 0.901 & 0.854 & 0.627 & 0.320 & 0.862 \\
        & 16384 & 50.7 & 0.712 & 0.844 & 0.861 & 0.661 & 0.338 & 0.855 \\
    \bottomrule
    \end{tabular}
    \label{tab:saebench}
\end{table*}

\subsection{SAE Feature Quality}

In this subsection, we evaluate HSAE from a standard SAE perspective by treating each level as an independent SAE.
We first examine the reconstruction–sparsity trade-off by measuring the average $L_0$ and the fraction of variance explained. Following the SAEBench evaluation protocol~\cite{Karvonen2025SAEBenchAC}, we further compare HSAE with baseline SAEs on a suite of established benchmarks, including Absorption (Absorp)~\cite{Chanin2024AIF}, Automated Interpretability (AutoInterp)~\cite{Paulo2024AutomaticallyIM}, Spurious Correlation Removal (SCR)~\cite{Karvonen2025SAEBenchAC}, Sparse Probing~\cite{gao2025scaling}, and RAVEL~\cite{Huang2024RAVELEI}. Additional evaluation details are provided in Appendix~\ref{app:exp:saebench}.

As shown in \Cref{tab:saebench}, HSAE achieves performance comparable to, and in some cases exceeding, baseline SAEs across most metrics. The overall results demonstrate that the hierarchical organization captured by HSAE does not come at the expense of feature quality. Instead, HSAE provides these structural insights as an additional benefit while maintaining the competitive reconstruction and interpretability standards of modern SAEs.

One notable result is the absorption score, where HSAE substantially outperforms baseline SAEs, particularly at larger dictionary sizes. While mitigating feature absorption is not an explicit goal of HSAE, this advantage is consistent with findings from Matryoshka SAE~\cite{bussmann2025Matryoshka}, which also induces a coarse-grained hierarchical ordering among features. Despite substantial differences in implementation between HSAE and Matryoshka SAE, the similarity in both design principles and empirical outcomes suggests that introducing hierarchical structure may help mitigate the feature absorption problem.

\subsection{Ablation Studies}

\begin{figure}[ht]
    \centering
    \includegraphics[width=0.7\columnwidth]{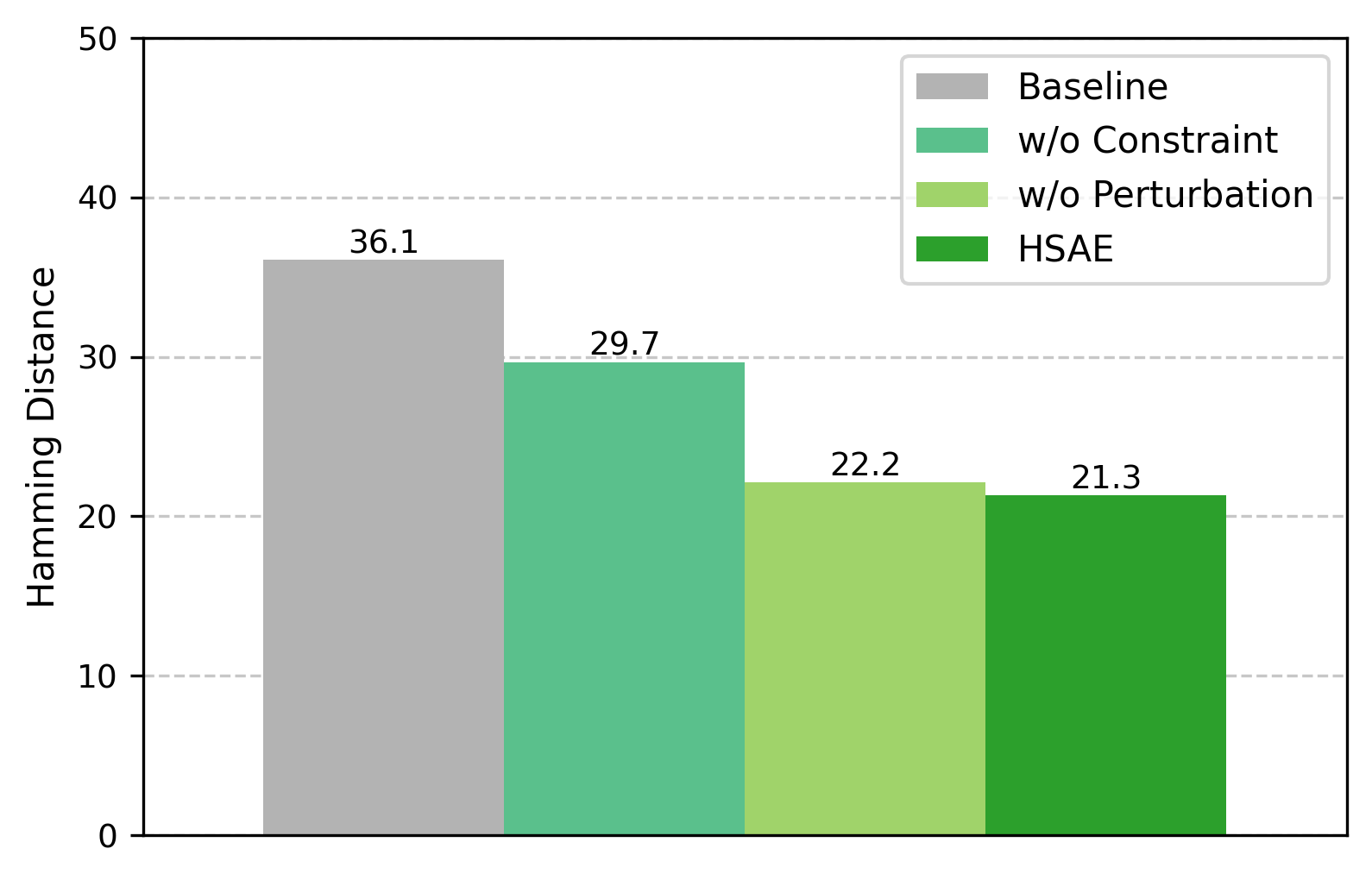}
    \caption{
    \textbf{Ablation of training mechanisms.} We compare the parent-child alignment across: (1) \textit{Baseline}: independently trained SAEs with post-hoc alignment; (2) \textit{w/o Constraint}: without parent-child constraint loss; (3) \textit{w/o Perturbation}: without random perturbation; and (4) \textit{HSAE}: our complete implementation.
    }
    \label{fig:ablation}
\end{figure}

To isolate the contributions of different components in HSAE, we investigate two primary mechanisms for enforcing hierarchical alignment: \textbf{the parent-child constraint loss} and \textbf{random feature perturbation}. We evaluate their efficacy by measuring parent-child alignment across several configurations as shown in \Cref{fig:ablation}.

The results demonstrate that the full HSAE achieves the lowest Hamming distance of 21.3, confirming that both mechanisms are essential for maintaining hierarchical consistency. Explicit constraint provides the foundational alignment, as its removal causes the Hamming distance to rise significantly to 29.7. Random perturbation further refines this relationship, reducing the distance by an additional 0.9 compared to using the constraint loss alone. Collectively, these mechanisms act complementarily to ensure that parent features faithfully encapsulate the collective activation patterns of their children.

We further investigate how different tree topology assumptions affect hierarchical alignment. Detailed results and comparative analyses are provided in Appendix~\ref{app:exp:more_exp}.

\subsection{More Observations}

In this section, we present further insights into the relationship between the learned hierarchical structure and the underlying activation space.

\begin{figure}[ht]
    \centering
    \includegraphics[width=0.7\columnwidth]{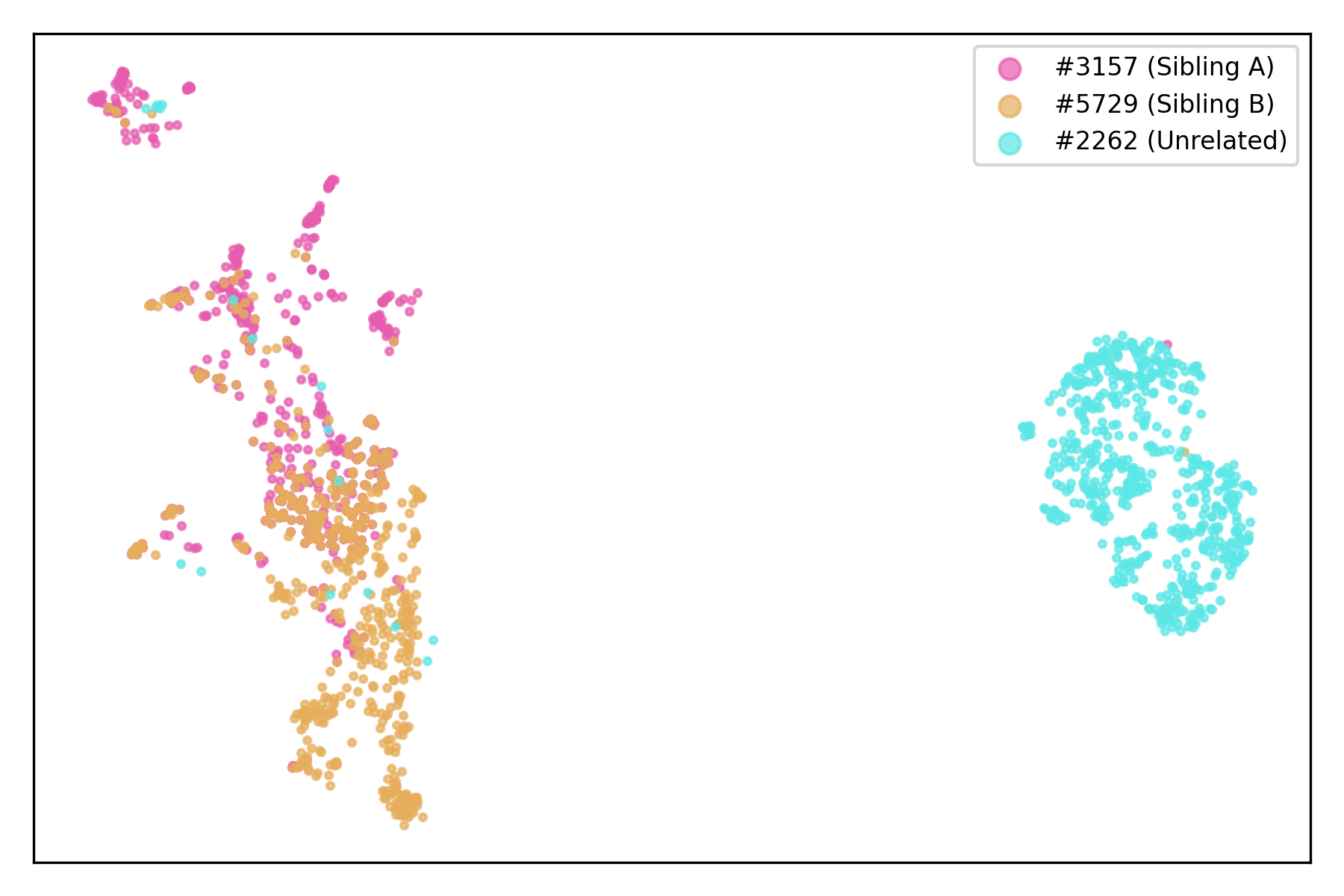}
    \caption{
    \textbf{Geometric manifestation of hierarchical relations.} UMAP projection of activations that trigger different features. Sibling A (\#3157) and Sibling B (\#5729) are features at level 2 that share the same parent (\#14).
    }
    \label{fig:activation_umap}
\end{figure}

\paragraph{Geometric Manifestation of Hierarchy.} We first examine whether the discovered hierarchical relationships are reflected in the geometry of the activation space. \Cref{fig:activation_umap} shows the UMAP projection of activations that trigger various features. We observe that activations for sibling features (those sharing a common parent) tend to cluster more closely together than those of unrelated features. This alignment indicates that the semantic hierarchy is not merely a structural constraint imposed by HSAE, but is also inherently represented in the geometric distribution of the LLM’s activations. To demonstrate that \Cref{fig:activation_umap} is representative of a broader trend, we provide additional examples in Appendix~\ref{app:exp:observations}.

\paragraph{Branching Factor and Semantic Clarity.} We further investigate how a feature's structural properties correlate with its interpretability. Specifically, we compare the AutoInterp scores of root features having only a single child against those that branch into multiple children. We find that multi-child roots exhibit a 2.49\% higher average score—a performance delta comparable to the interpretability gain typically achieved by quadrupling the dictionary size (from 16k to 64k) in standard SAEs. This suggests an underlying link between a feature's hierarchical positioning and the semantic clarity of its represented concept. More detailed results are provided in \Cref{fig:autointerp_gap} in Appendix~\ref{app:exp:observations}.

\section{Conclusion and Limitation}

We introduced the Hierarchical Sparse Autoencoder, the first framework designed to explicitly learn conceptual hierarchies alongside sparse features. Through a joint optimization objective and a random perturbation mechanism, HSAE recovers meaningful taxonomies that align with human intuition. Our experiments demonstrate that HSAE significantly outperforms post-hoc alignment baselines in hierarchical consistency while preserving the feature quality of standard SAEs. Furthermore, we report unique observations linking the learned structure to the geometric properties of the activation space and feature interpretability.


Despite these advancements, several limitations remain. First, the current tree structure is constrained by a fixed number of levels, which may not fully capture the varying depths of semantic abstraction in large-scale models. Second, although parent-child co-activation is significantly improved, some misalignments persist; it remains unclear if these reflect methodological constraints or the inherently non-hierarchical nature of certain activations. Future work should investigate investigate more flexible architectural priors and the scaling laws of hierarchical discovery. Beyond methodology refinements, these hierarchies provide a foundation for multi-scale model steering and safety auditing, enabling researchers to intervene in model behavior at the specific level of semantic granularity.

\section*{Impact Statement}

This paper presents work whose goal is to advance the field of machine learning by improving the interpretability of LLMs. By providing tools to uncover hierarchical structures in latent representations, our work contributes to the broader effort of making AI systems more transparent and understandable. There are many potential societal consequences of our work, none of which we feel must be specifically highlighted here.

\section*{Acknowledgment}
Bin Dong is supported in part by the New Cornerstone Investigator Program. Zhennan Zhou is supported by Zhejiang Provincial Natural Science Foundation of China, Project Number QKWL25A0501, and Fundamental and Interdisciplinary Disciplines Breakthrough Plan of the Ministry of Education of China, Project Number JYB2025XDXM502.

\bibliography{references}
\bibliographystyle{icml2026}

\newpage
\appendix

\onecolumn

\section{Detailed Settings}
\label{app:exp:setup}

\subsection{Experiment Setups}

\paragraph{Data Preparation.}
For all experiments, we collect a dataset of 100M activation vectors from the target LLM. These activations are generated using input prompts with a sequence length of 1024 tokens. Following standard practice, we exclude activations from the \texttt{bos} token position. All activation vectors are normalized with a constant scaler to have a RMS norm expectation of $\sqrt{d}$ before being used for training.

\paragraph{Training Hyperparameters.}
We train all models using a batch size of $1024$ with the Adam optimizer ($\beta_1=0, \beta_2=0.995$). We employ a cosine learning rate scheduler, starting with a linear warm-up phase covering the first $10\%$ of training steps, followed by a decay to zero.

\paragraph{Computational Resources and Efficiency.}
All training and evaluation tasks are conducted on NVIDIA A800 GPUs. For a target model with hidden dimension $d=2304$, training a 4-level HSAE (with dictionary sizes ranging from $2048$ to $16384$) takes approximately $6$ hours. Compared to training independent SAEs of equivalent sizes, HSAE introduces a modest computational overhead. This is primarily attributed to the calculation of the hierarchical constraint loss and the memory latency incurred by fragmented indexing during the feature perturbation.

\subsection{Dynamic Sparsity Control}

To maintain a consistent level of sparsity throughout training, we implement a feedback control mechanism to dynamically adjust the sparsity penalty weight $\lambda_\ell$. Specifically, we define a target sparsity $\hat{L}_0$ and model the desired trajectory of the average $L_0$ as a first-order exponential decay process:$$\Delta L_0 = -\eta (L_0 - \hat{L}_0)$$where $\eta$ represents the decay rate, set to $0.001$ in our experiments. We track the current $L_0$ using an exponential moving average (EMA) with a momentum of $0.999$ to filter out high-frequency noise from individual batches.During each training step, we compare the observed change in $L_0$ against the theoretical $\Delta L_0$ derived from the exponential model. If the observed reduction in sparsity is more rapid than the target trajectory, the sparsity weight $\lambda_\ell$ is decreased; conversely, if the model remains denser than expected, $\lambda_\ell$ is increased. In practice, we found that this velocity-based control significantly reduces oscillations in training compared to the control methods described in \citet{Karvonen2025SAEBenchAC}.

\section{More Case Studies}
\label{app:exp:more_empirical}

In this section, we provide further qualitative evidence of the hierarchical structures discovered by HSAE.

\begin{figure}[ht]
    \centering
    \begin{subfigure}[b]{0.4\textwidth}
        \centering
        \includegraphics[width=\textwidth]{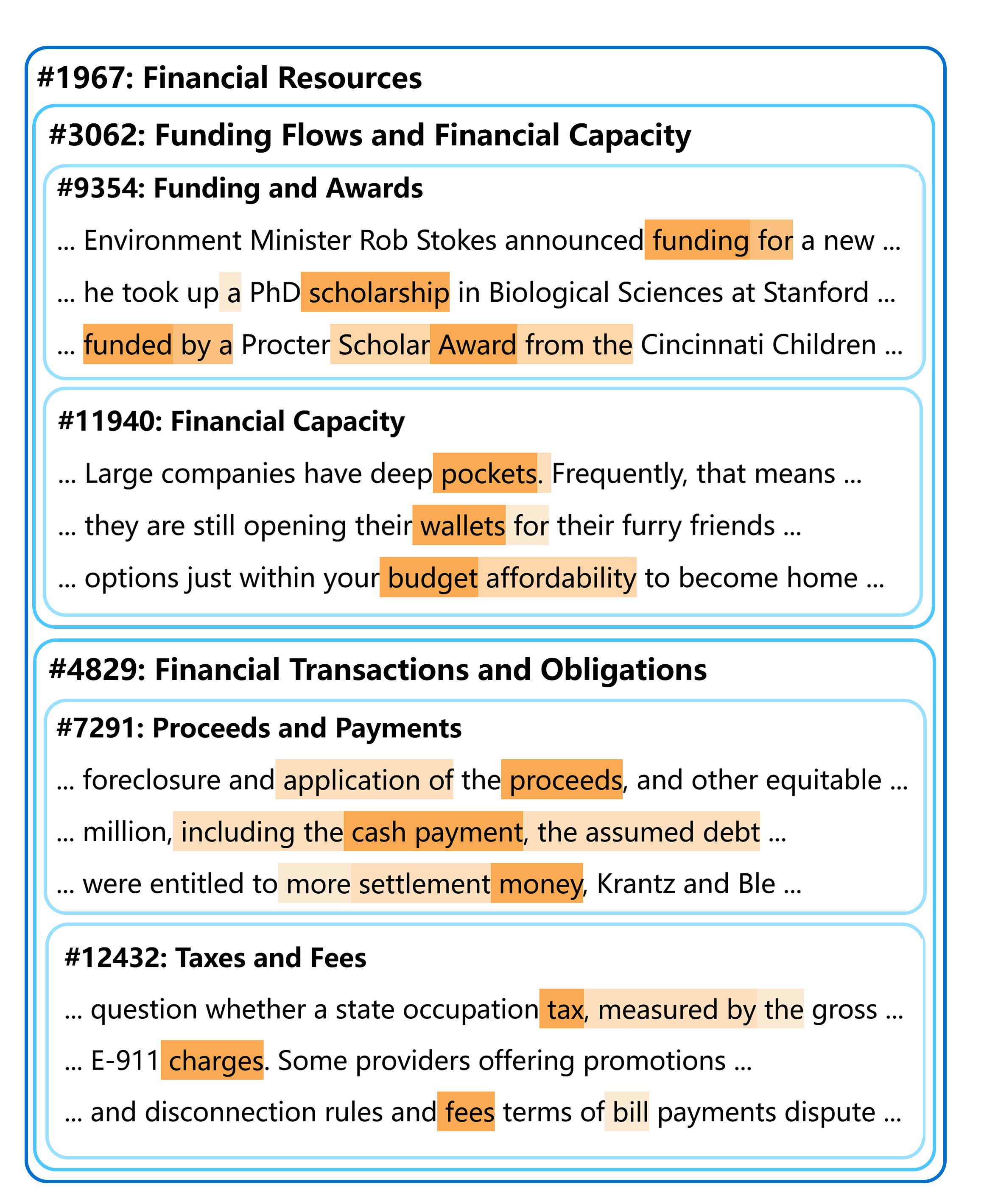}
        \caption{
        \textbf{Hierarchical feature discovery of financial resource concepts.} The root feature (\#1967) representing financial concepts branches into \textit{Funding Flows and Financial Capacity} (\#3062) and \textit{Financial Transactions and Obligations} (\#4829). Within the \textit{Funding Flows} branch, HSAE further distinguishes between external support (\textit{Funding and Awards}, \#9354) and internal economic status (\textit{Financial Capacity}, \#11940). Within the latter, the hierarchy resolves into granular features for specific settlements (\textit{Proceeds and Payments}, \#7291) and statutory costs (\textit{Taxes and Fees}, \#12432).
        }
        \label{fig:empirical_3}
    \end{subfigure}
    \qquad
    \begin{subfigure}[b]{0.4\textwidth}
        \centering
        \includegraphics[width=\textwidth]{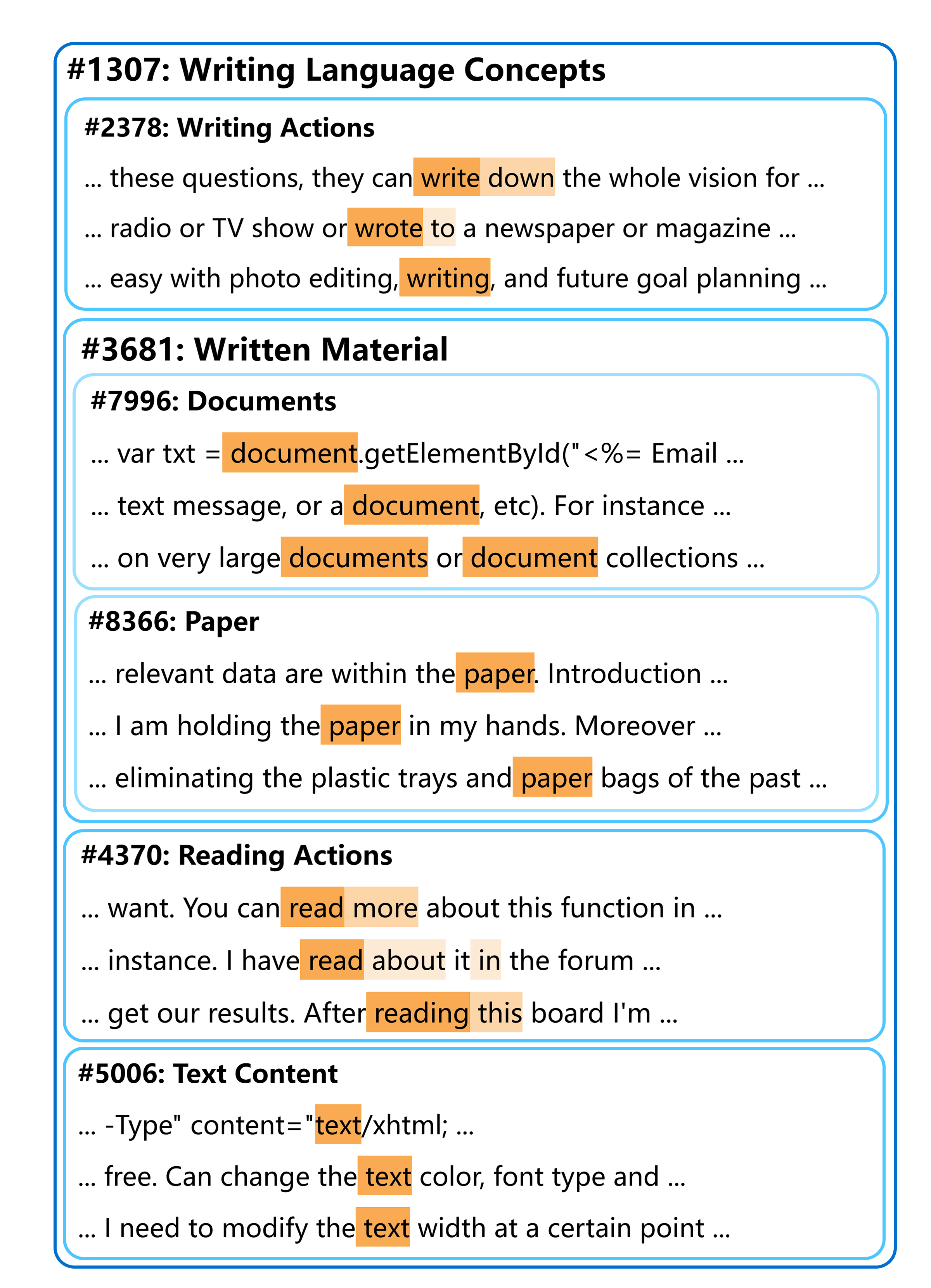}
        \caption{
        \textbf{Hierarchical feature discovery of writing and language concepts.} The root feature (\#1307) representing broad language-related concepts branches into specialized domains: \textit{Writing Actions} (\#2378), \textit{Written Material} (\#3681), \textit{Reading Actions} (\#4370), and \textit{Text Content} (\#5006). Within the \textit{Written Material} branch, HSAE further distinguishes between digital or structured documents (\textit{Documents}, \#7996) and physical or raw media (\textit{Paper}, \#8366).
        }
        \label{fig:empirical_2}
    \end{subfigure}
    \caption{\textbf{Extended Case Studies of Hierarchical Feature Discovery.}}
    \label{fig:extra_empirical}
\end{figure}

Figure~\ref{fig:extra_empirical} (a) illustrates the hierarchical organization of \textit{Financial Resource Concepts}. A broad root feature (\#1967) serves as the semantic anchor, which the model partitions into distinct functional domains: capital acquisition and fiscal status, categorized as \textit{Funding Flows and Financial Capacity} (\#3062), and the execution of financial duties, labeled as \textit{Financial Transactions and Obligations} (\#4829). Within these branches, the hierarchy further resolves into specialized leaf features that capture subtle semantic nuances. For instance, the model distinguishes between formal institutional support (e.g., \textit{scholarship}, \textit{awards} in \#9354) and informal or abstract expressions of wealth (e.g., \textit{deep pockets}, \textit{budget} in \#11940). Concurrently, the transactional branch successfully separates commercial settlements (e.g., \textit{proceeds}, \textit{cash payment} in \#7291) from mandatory statutory costs (e.g., \textit{tax}, \textit{fees} in \#12432).

Figure~\ref{fig:extra_empirical} (b) illustrates the hierarchical organization of \textit{Writing and Language Concepts}. A broad root feature (\#1307) serves as the semantic anchor, which the model partitions into distinct functional domains: active processes such as \textit{Writing} (\#2378) and \textit{Reading Actions} (\#4370), and their corresponding outputs, categorized as \textit{Text Content} (\#5006) and \textit{Written Material} (\#3681). Within the \textit{Written Material} branch, the hierarchy further resolves into granular sub-features that distinguish between structured digital formats (\textit{document}, \#7996) and physical media (\textit{paper}, \#8366).

\section{Auto-Interpretability of Hierarchies}
\label{app:exp:autointerp}
This appendix provides implementation details for the auto-interpretability assessment described in Section 4.3. Our pipeline extends the SAEBench evaluation framework \citep{Karvonen2025SAEBenchAC} with custom modifications to evaluate the hierarchical relations between HSAE features.

\paragraph{Feature Pairs Sampling and Data Collection.} We randomly sample 5000 parent-child pairs from the learned hierarchical structure in HSAE. Activations are computed on sequences from the mini-PILE testing dataset. Following the SAEBench methodology, for each feature we collect 10 sequences with the highest activation values and sample 5 sequences with probability proportional to their activation values. These sequences are formatted by highlighting activating tokens with \texttt{<<token>>} syntax.

\paragraph{LLM Judge Configurations.}  We employ \texttt{Qwen3-Max} as the judge model. 

The system prompt is as follows:

\texttt{We're studying neurons in a sparse autoencoder (SAE) within a neural network. Each neuron activates on specific words, substrings, or concepts in short documents, with activating words indicated by << ... >>. You will be given two sets of documents where two different neurons activate. Your task is to compare the activation patterns of these two neurons and determine if there is a parent-child relationship between them. A parent-child relationship means that one neuron's activating concept (the child) is a subset or a more specific version of the other neuron's activating concept (the parent). Analyze the provided examples and output your judgment in the following format:\\
\\
HaveRelationship: [Yes/No]\\
Confidence: [High/Medium/Low]\\
\\
Do not include any additional text, explanations, or formatting.
}

The user prompt template:

\texttt{Here are the activating documents for Neuron A:
\newline
[list of examples with activating tokens highlighted]\\
\\
And for Neuron B:
\newline
[list of examples with activating tokens highlighted]\\
\\
Based on these documents, determine if Neuron A and Neuron B have a parent-child relationship.
}

\section{SAEBench Evaluation Details}
\label{app:exp:saebench}

We evaluate our models using the SAEBench framework~\cite{Karvonen2025SAEBenchAC}, adopting the default configurations for comparability. Since SAEBench provides a diverse set of metrics, we specify the exact metric identifiers used for our reporting in \Cref{tab:detail_metric}. Notably, for Absorption, we report ($1 - \text{score}$) to align it with other metrics where a higher value indicates superior performance.

\begin{table}[ht]
\centering
\caption{Mapping of reported benchmarks to specific SAEBench metrics.}
\label{tab:detail_metric}
\begin{tabular}{|c|c|}
    \hline
        \textbf{Benchmark} & \textbf{Metric} \\
    \hline
        Variance Explained & \texttt{explained\_variance\_legacy} \\
    \hline
        Absorption & \texttt{mean\_absorption\_fraction\_score}*\\
    \hline
        AutoInterp & \texttt{autointerp\_score} \\
    \hline
        RAVEL & \texttt{disentanglement\_score} \\
    \hline
        SCR & \texttt{scr\_metric\_threshold\_20} \\
    \hline
        Sparse Probing & \texttt{sae\_top\_5\_test\_accuracy} \\
    \hline
\end{tabular}
\end{table}

\section{More activation sources and target LLMs}
\label{app:exp:more_exp}

To evaluate the generalizability of HSAE, we extend our experiments across multiple dimensions. First, we move beyond the 13th layer of \texttt{gemma-2-2b} to train HSAE on residual stream activations from early and late stages (layers 6 and 20). Second, we apply our method to the 18th layer of \texttt{qwen3-4b}~\cite{Yang2025Qwen3TR} to verify cross-model consistency. Finally, we test the robustness of the discovered hierarchy under varying capacity constraints by training on \texttt{gemma-2-2b-layer-13} with target sparsities $L_0 \in \{80, 100\}$.

As summarized in \Cref{tab:more_exp}, HSAE consistently maintains reconstruction fidelity comparable to the baselines across all tested settings. Here, the reported $L_0$ and Variance Explained represent the average performance across all hierarchical levels. Simultaneously, it yields significant improvements in parent-child alignment and co-activation probabilities. These results collectively demonstrate HSAE's ability to recover structured feature taxonomies is not dependent on specific layer depths, model architectures, or sparsity levels.

We note a slight discrepancy in the \textit{Variance Explained} scores for \texttt{gemma2-2b-layer-13} between \Cref{tab:saebench} and \Cref{tab:more_exp}. This shift is attributed to the different evaluation corpora employed: the benchmarks in \Cref{tab:saebench} strictly follow the SAEBench protocol using OpenWebText, whereas the results in \Cref{tab:more_exp} are computed using the mini-PILE testing dataset. Despite these distributional shifts, the reconstruction fidelity between HSAE and baseline remain comparable.

\begin{table*}[ht]
    \centering
    \caption{Evluation results for HSAE and baseline SAEs on multiple benchmarks. All metrics except $L_0$ are the higher the better.}
    \begin{tabular}{ccccccc}
    \toprule
        \textbf{Activation Source} & \textbf{Model} & \textbf{$L_0$} & \textbf{Var. Exp.} & $P(\text{activ}_p \mid \text{activ}_c)$ & $P(\text{activ}_c \mid \text{activ}_p)$ & \textbf{Ham. Dist. ($\downarrow$)} \\
    \midrule
        \multirow{2}{*}{\texttt{gemma2-2B-layer-6}} 
        & HSAE      & 50.1 & 0.759 & 0.752 & 0.403 & 23.2 \\
        & Baseline  & 50.0 & 0.759 & 0.651 & 0.331 & 34.6 \\
    \midrule
        \multirow{6}{*}{\texttt{gemma2-2b-layer-13}} 
        & \multirow{3}{*}{HSAE}      
          & 50.1 & 0.713 & 0.769 & 0.432 & 21.3 \\
        & & 80.1 & 0.743 & 0.749 & 0.421 & 36.2 \\
        & & 100.7 & 0.757 & 0.747 & 0.415 & 46.3 \\
        & \multirow{3}{*}{Baseline}  
          & 49.8 & 0.714 & 0.662 & 0.343 & 36.1 \\
        & & 80.0 & 0.743 & 0.634 & 0.326 & 61.9 \\
        & & 100.0 & 0.758 & 0.619 & 0.317 & 78.9 \\
    \midrule
        \multirow{2}{*}{\texttt{gemma2-2b-layer-20}} 
        & HSAE      & 50.0 & 0.729 & 0.800 & 0.461 & 19.4 \\
        & Baseline  & 50.0 & 0.727 & 0.729 & 0.395 & 30.6 \\
    \midrule
        \multirow{2}{*}{\texttt{qwen3-4b-layer-18}} 
        & HSAE      & 50.2 & 0.988 & 0.347 & 0.197 & 28.7 \\
        & Baseline  & 50.0 & 0.988 & 0.309 & 0.179 & 46.4 \\
    \bottomrule
    \end{tabular}
    \label{tab:more_exp}
\end{table*}

\section{Ablation Studies}
\label{app:exp:ablation}

\subsection{Similarity Metric for Hierarchy Updates}

\begin{figure}[ht]
    \centering
    \includegraphics[width=0.5\columnwidth]{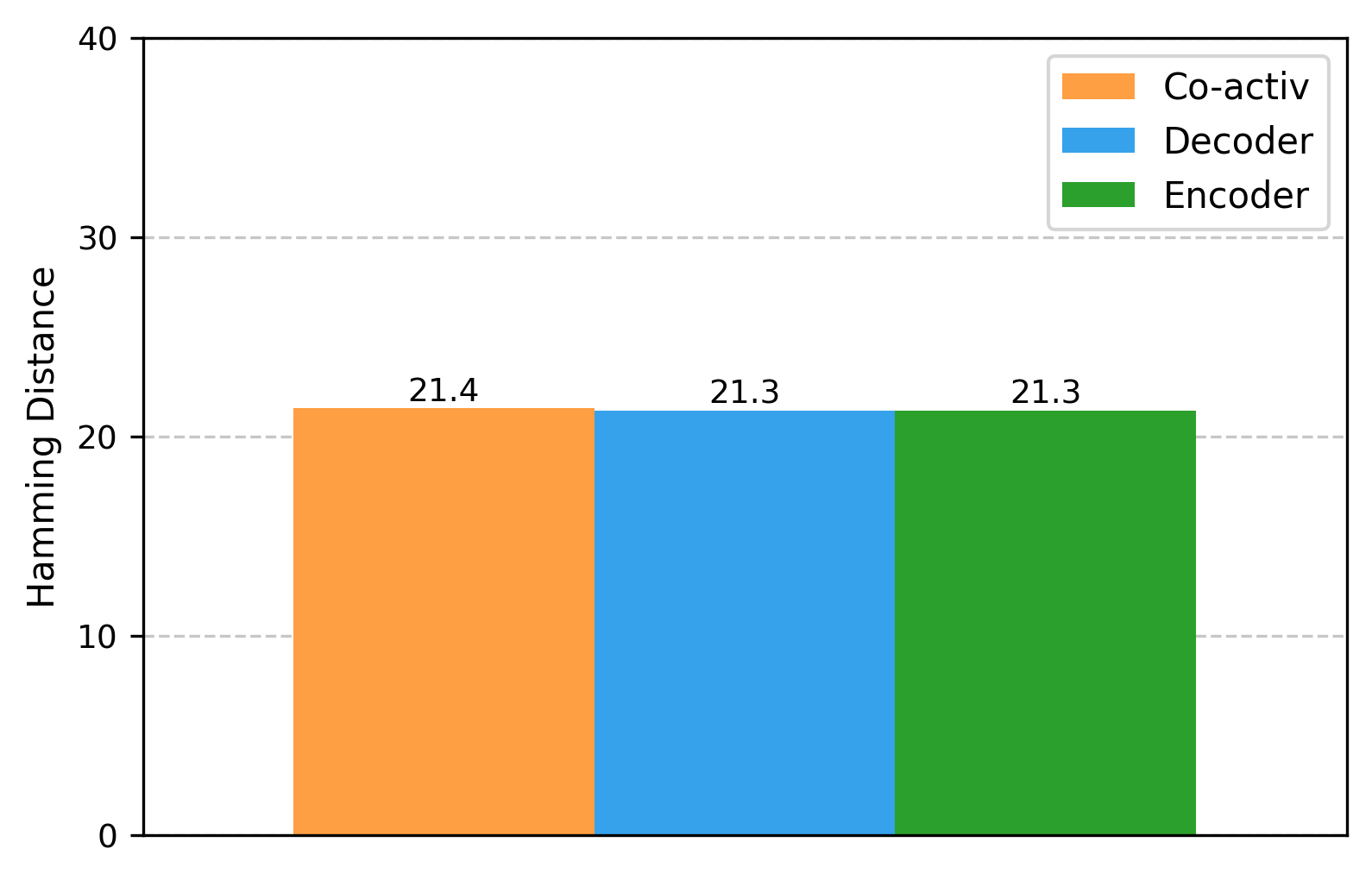}
    \caption{
    \textbf{Ablation of similarity metric used for hierarchy Update.} (1) \textit{Co-activ}: Use statistically derived co-activation probability as similarity; (2) \textit{Decoder}: Use decoder vector similarity; (3) \textit{Encoder}: Use encoder vector similarity. All three metrics provide similar results.
    }
    \label{fig:similarity_method}
\end{figure}

We investigate the impact of different similarity metrics used during the hierarchy update. As shown in \Cref{fig:similarity_method}, we compare three potential proxies for feature relatedness: (1) \textbf{Co-activation} (statistically derived co-activation probability), (2) \textbf{Decoder} (cosine similarity between decoder vectors), and (3) \textbf{Encoder} (cosine similarity between encoder vectors).

Our results indicate that all three metrics yield nearly indistinguishable performance in terms of the final hierarchical consistency. Since the choice of similarity metric does not significantly alter the outcome, we adopt Encoder similarity as the default for all experiments, as it offers the most direct and computationally efficient implementation within our training pipeline.

\subsection{Hyper-parameter Tuning}

We examine the impact of key hyper-parameters on HSAE performance, specifically the parent-child constraint weight $\rho$ and the random perturbation rate $r$. These parameters are essential for effective hierarchical training; as expected, setting both to zero would cause the method to degrade into independent SAEs.

\begin{figure*}[ht]
    \centering
    \begin{subfigure}[b]{0.4\textwidth}
        \centering
        \includegraphics[width=\textwidth]{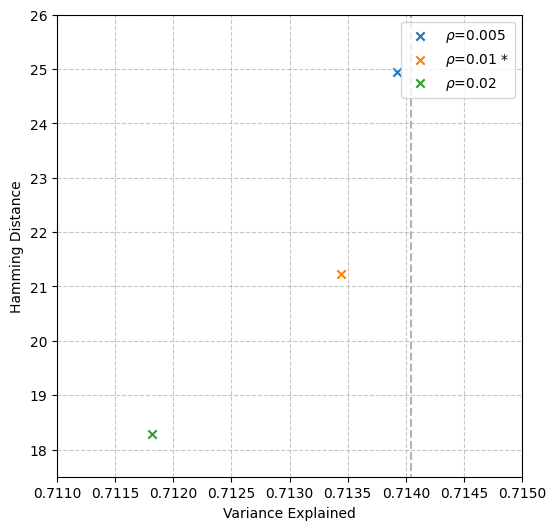}
        \caption{Parent-Child Constraint Weight $\rho$}
    \end{subfigure}
    \begin{subfigure}[b]{0.4\textwidth}
        \centering
        \includegraphics[width=\textwidth]{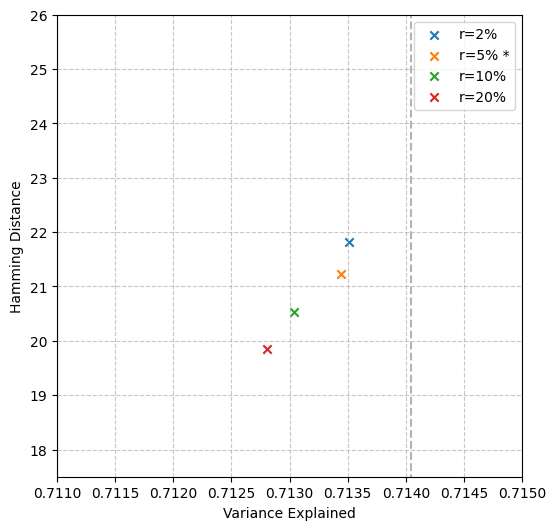}
        \caption{Random Perturbation Rate $r$}
    \end{subfigure}
    \caption{
    \textbf{Hyper-parameters' effects to reconstruction-hierarchy consistency trade-off.}
    Increasing parent-child constraint weight $\rho$ or perturbation rates $r$ improve hierarchical alignment but eventually degrades Variance Explained.
    }
    \label{fig:hyperpara}
\end{figure*}

Our analysis reveals a nuanced trade-off: while the hierarchical constraints guide organized feature discovery, an excessively large constraint weight $\rho$ imposes overly rigid structural priors that can bottleneck the representational capacity of the features, leading to a decrease in Variance Explained. Similarly, a perturbation rate $r$ that is too high introduces excessive noise into the training dynamics, potentially causing instability and degraded feature quality. Unlike the ablations discussed in the main text—where all settings maintain high fidelity—these extreme configurations demonstrate a non-negligible trade-off between hierarchy consistency and reconstruction performance. We utilize the scatter plots in \Cref{fig:hyperpara} to visualize these relationships across different parameter regimes.

The final selection of hyper-parameters represents a principled balance between reconstruction fidelity and structural hierarchy consistency. We prioritize configurations that preserve maximal fidelity—ensuring the SAE effectively captures the original activation space—while simultaneously achieving optimal hierarchical alignment.

\subsection{Tree Structural Assumptions}

\begin{figure}[ht]
    \centering
    \includegraphics[width=0.5\columnwidth]{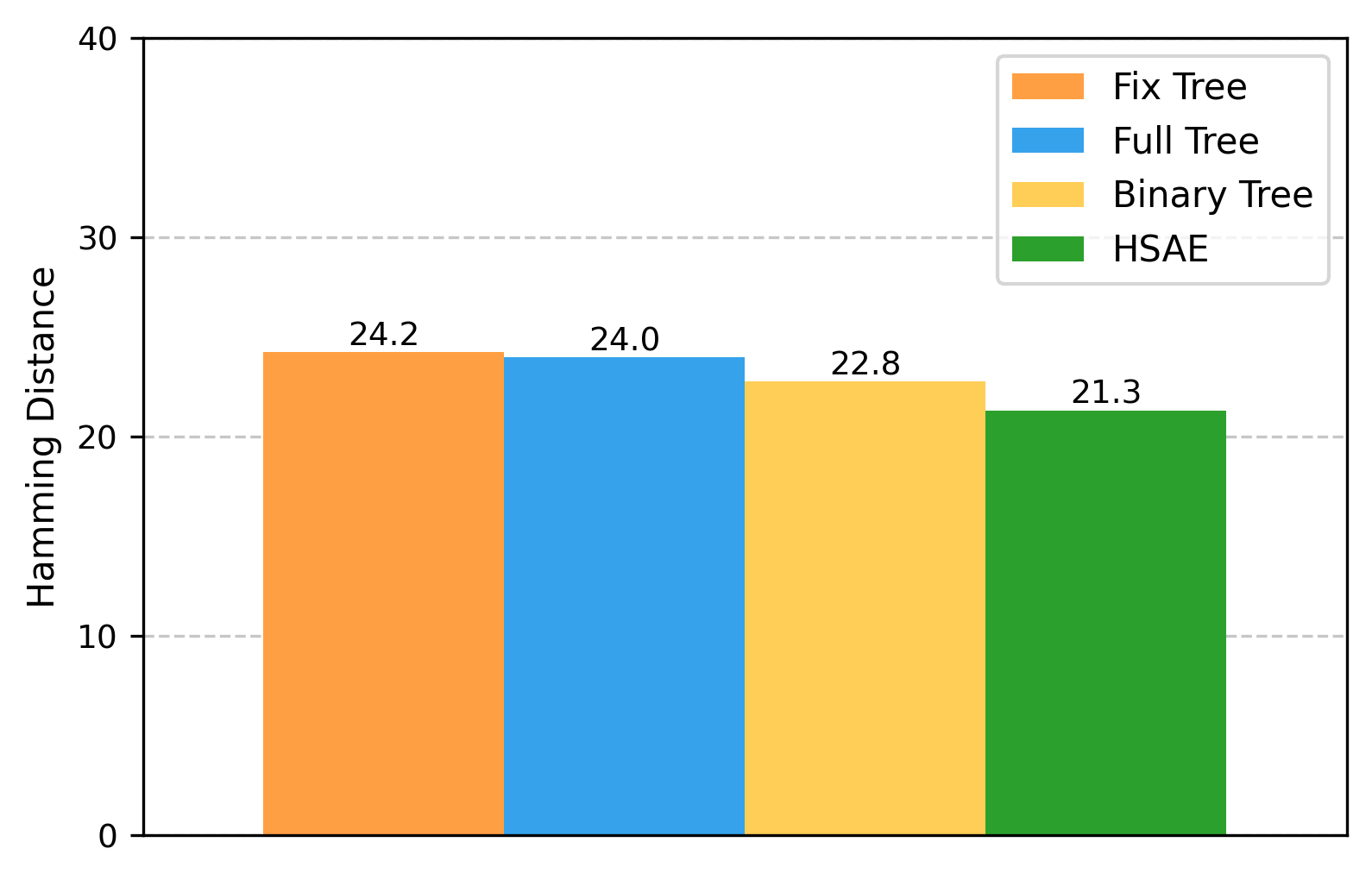}
    \caption{
    \textbf{Ablation of tree structural assumptions.} We compare parent-child alignment across different structural variants: (1) \textit{Fix Tree}: static hierarchy after early initialization; (2) \textit{Full Tree}: mandatory assignment for all features; (3) \textit{Binary Tree}: a hierarchy constrained to a maximum branching factor of two; and (4) \textit{HSAE}: our flexible partial tree approach.
    }
    \label{fig:ablation_tree}
\end{figure}

We further investigate how different architectural assumptions regarding the tree topology affect hierarchical alignment. Specifically, we compare the default HSAE with three alternative configurations: \textbf{Binary Tree}, which restricts each parent to a maximum of two children; \textbf{Fix Tree}, where the hierarchy is determined early in training and remains static; and \textbf{Full Tree}, which forcefully assigns every feature to a parent.

As shown in \Cref{fig:ablation_tree}, the default HSAE outperforms all constrained variants. The performance gap compared to the Binary Tree value of 22.8 suggests that conceptual decomposition in LLMs is naturally multi-branching rather than strictly dyadic. The significantly higher Hamming distance of the Fix Tree at 24.2 highlights the necessity of an alternating optimization process, as feature representations and their optimal hierarchical assignments must co-evolve during training. Finally, the Full Tree baseline yields sub-optimal consistency with a distance of 24.0. This indicates that the more flexible Partial Tree structure is necessary for capturing a more robust hierarchical structure of LLM activations, as it avoids forcing assignments for features that do not exhibit strong similarity to any potential parent.

\section{More Observations}

\label{app:exp:observations}

\subsection{Geometric Manifestation of Hierarchy}

\begin{figure*}[ht]
    \centering
    \begin{subfigure}[b]{0.4\textwidth}
        \centering
        \includegraphics[width=\textwidth]{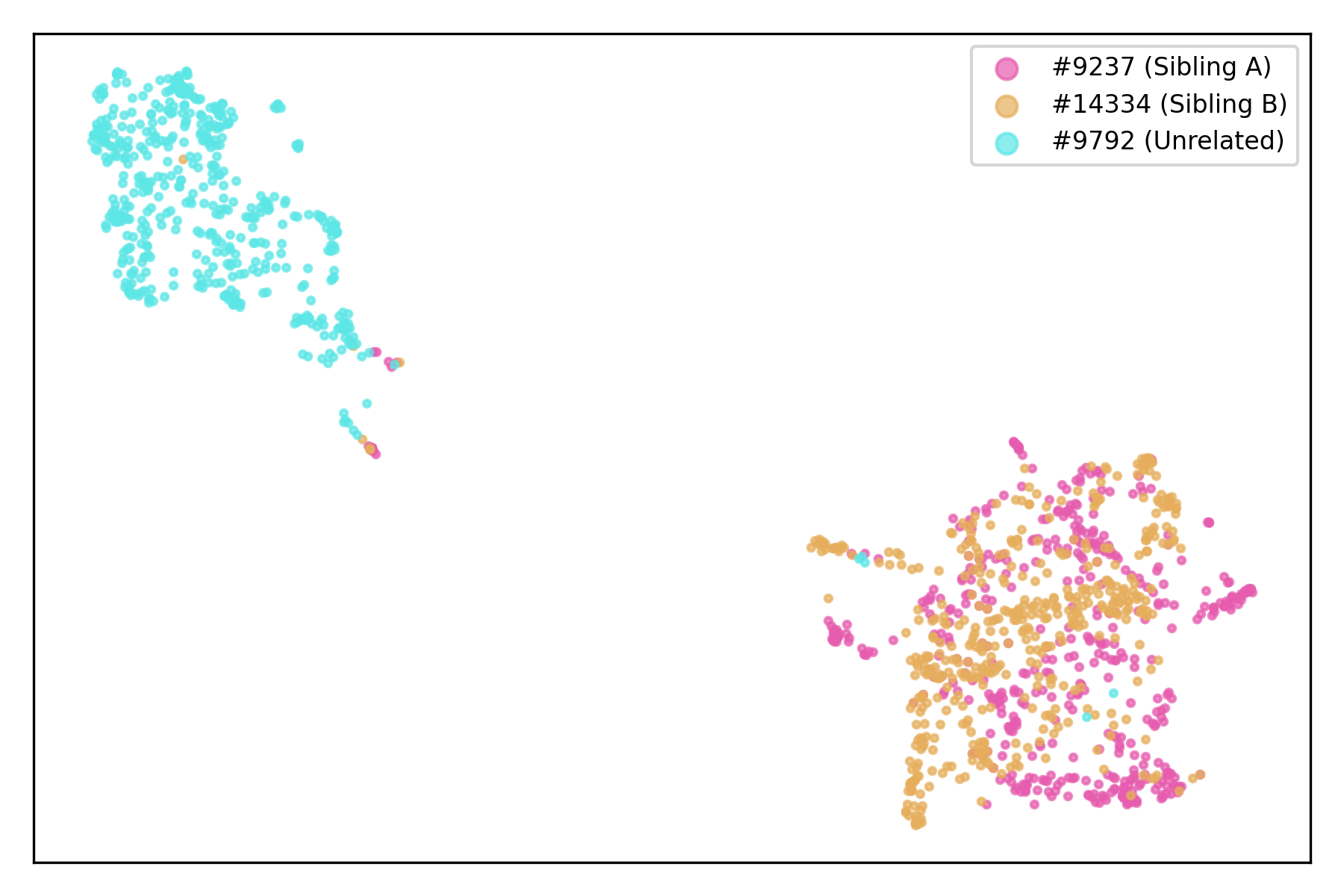}
    \end{subfigure}
    \begin{subfigure}[b]{0.4\textwidth}
        \centering
        \includegraphics[width=\textwidth]{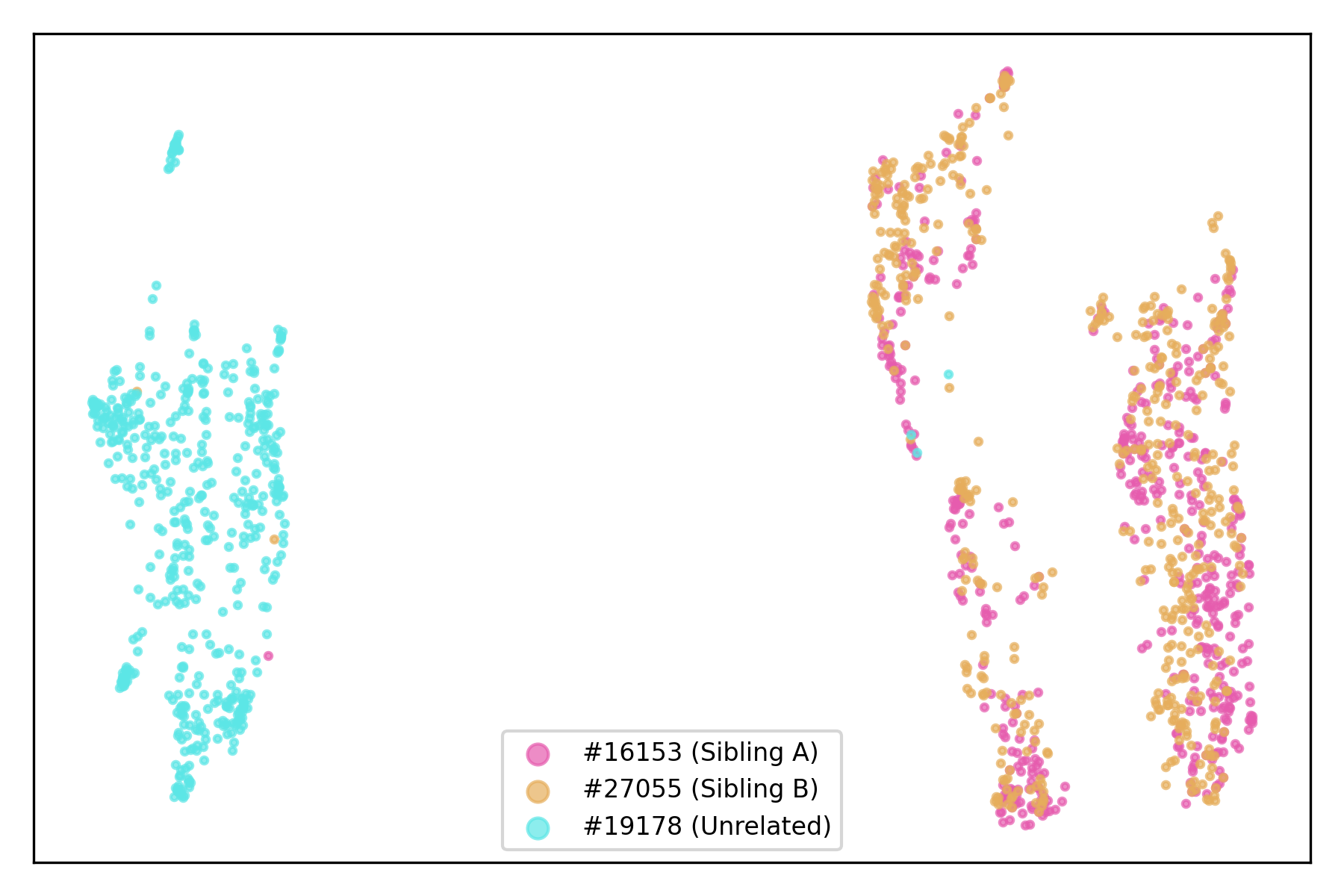}
    \end{subfigure}\\
    \begin{subfigure}[b]{0.4\textwidth}
        \centering
        \includegraphics[width=\textwidth]{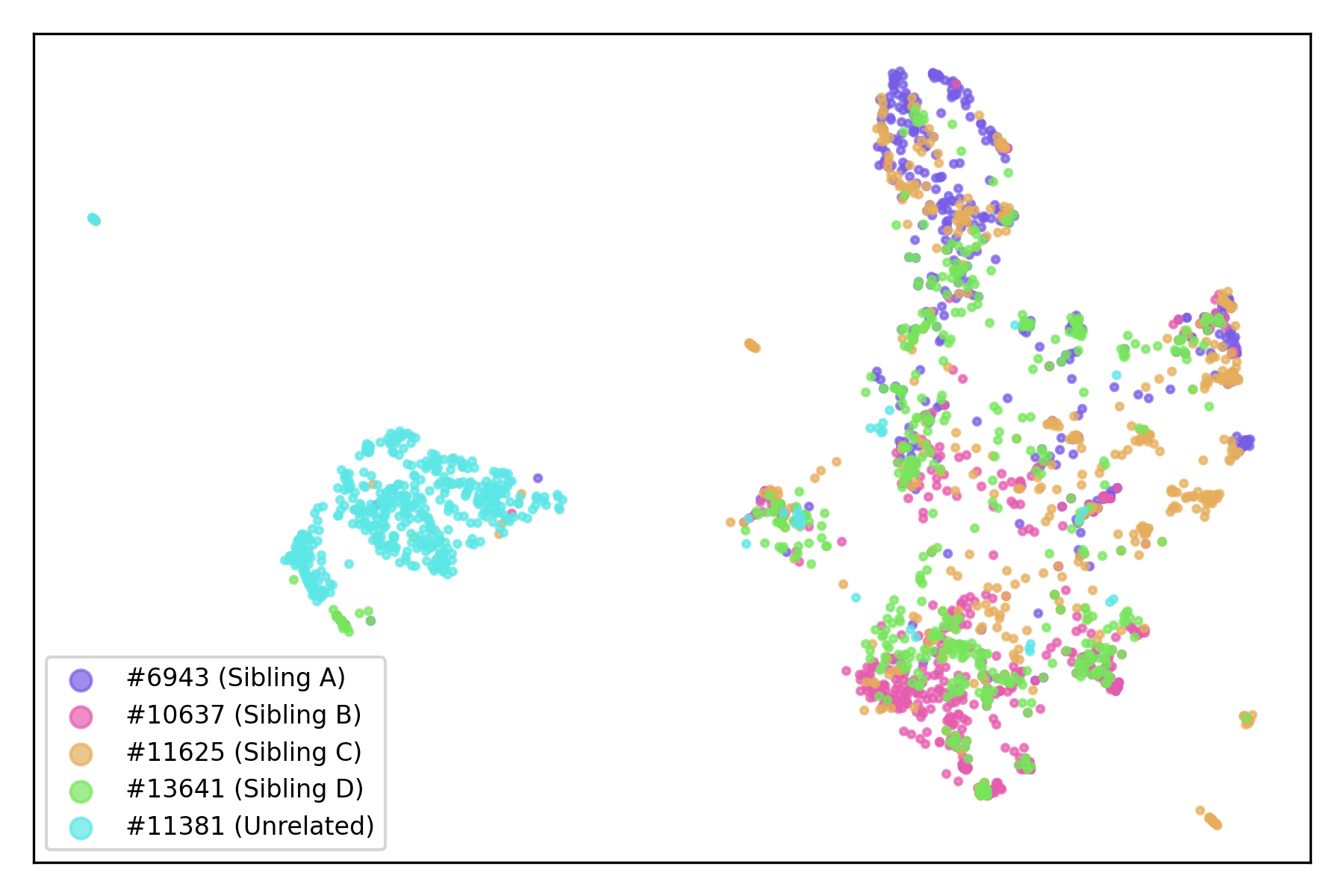}
    \end{subfigure}
    \begin{subfigure}[b]{0.4\textwidth}
        \centering
        \includegraphics[width=\textwidth]{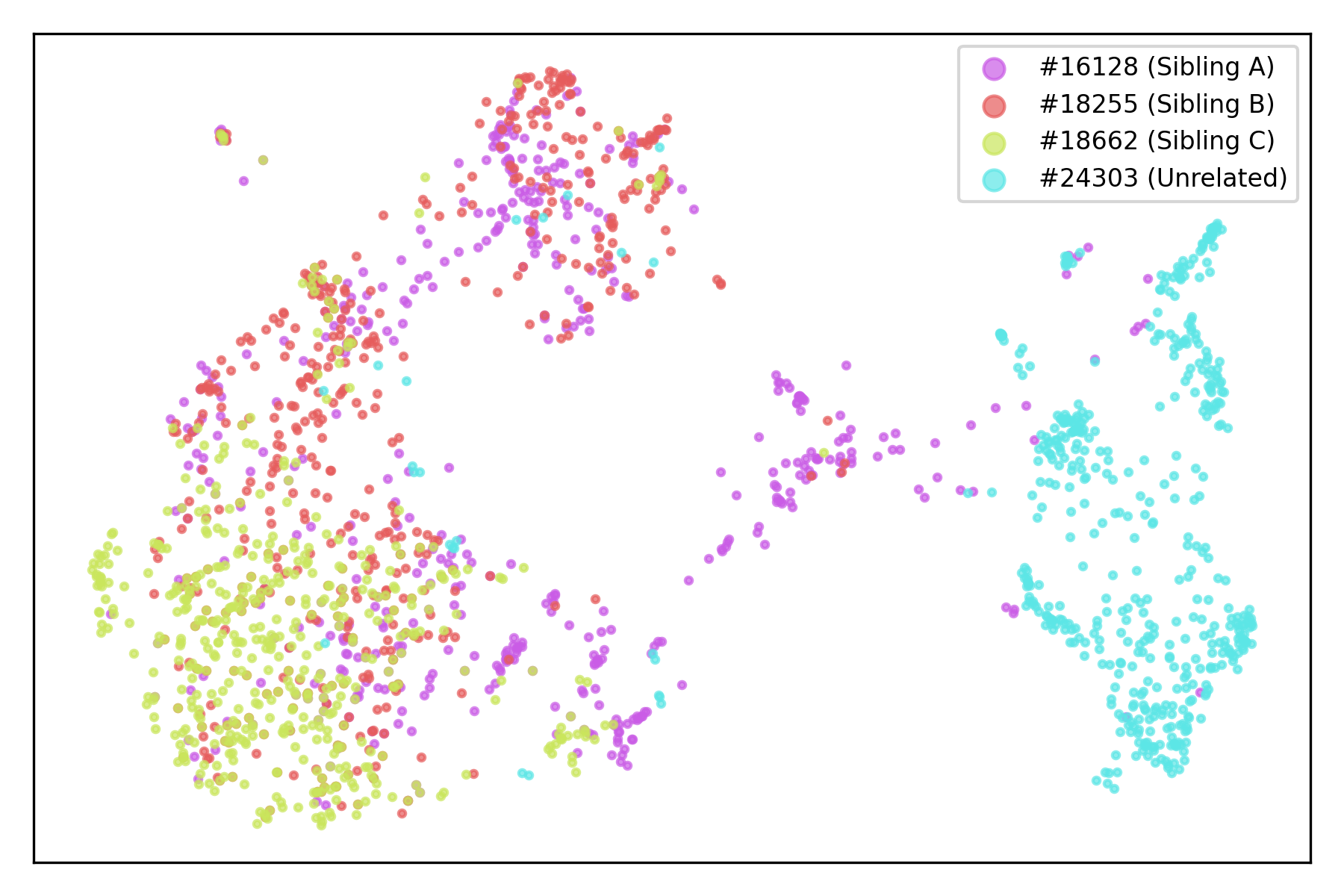}
    \end{subfigure}
    \caption{
    \textbf{Geometric manifestation of hierarchical relations.} UMAP projection of activations that trigger different features. In all cases, the unrelated feature is clearly seperated from sibiling features.
    }
    \label{fig:extra_umap}
\end{figure*}

To understand how the discovered hierarchy manifests in the original activation space, we visualize the activation clusters using UMAP projection in \Cref{fig:extra_umap}. Each plot contrasts the activations triggered by a set of sibling features (which share the same parent) against those triggered by an unrelated feature from a distant branch within the same hierarchical level.

We observe that unrelated features remain clearly isolated from sibling clusters. This geometric arrangement suggests that HSAE's structural constraints successfully capture the natural clustering of the LLM's latent space, where features with a common hierarchical ancestor are encoded in nearby geometric regions. 

While sibling features exhibit spatial proximity in the UMAP visualization, we should note that their co-activation rate does not increase compared to baseline SAEs. This indicates that HSAE learns a disentangled decomposition rather than simply clustering redundant features. Specifically, although the parent-child constraint encourages structural organization, it does not force sibling features to fire simultaneously on the same tokens. On the contrary, the sparsity penalty inherent in the SAE objective naturally discourages such co-activation, ensuring that each sibling feature remains a distinct and sparse functional unit.

\subsection{Branching Factor and Semantic Clarity}

\begin{figure}[ht]
    \centering
    \includegraphics[width=0.5\columnwidth]{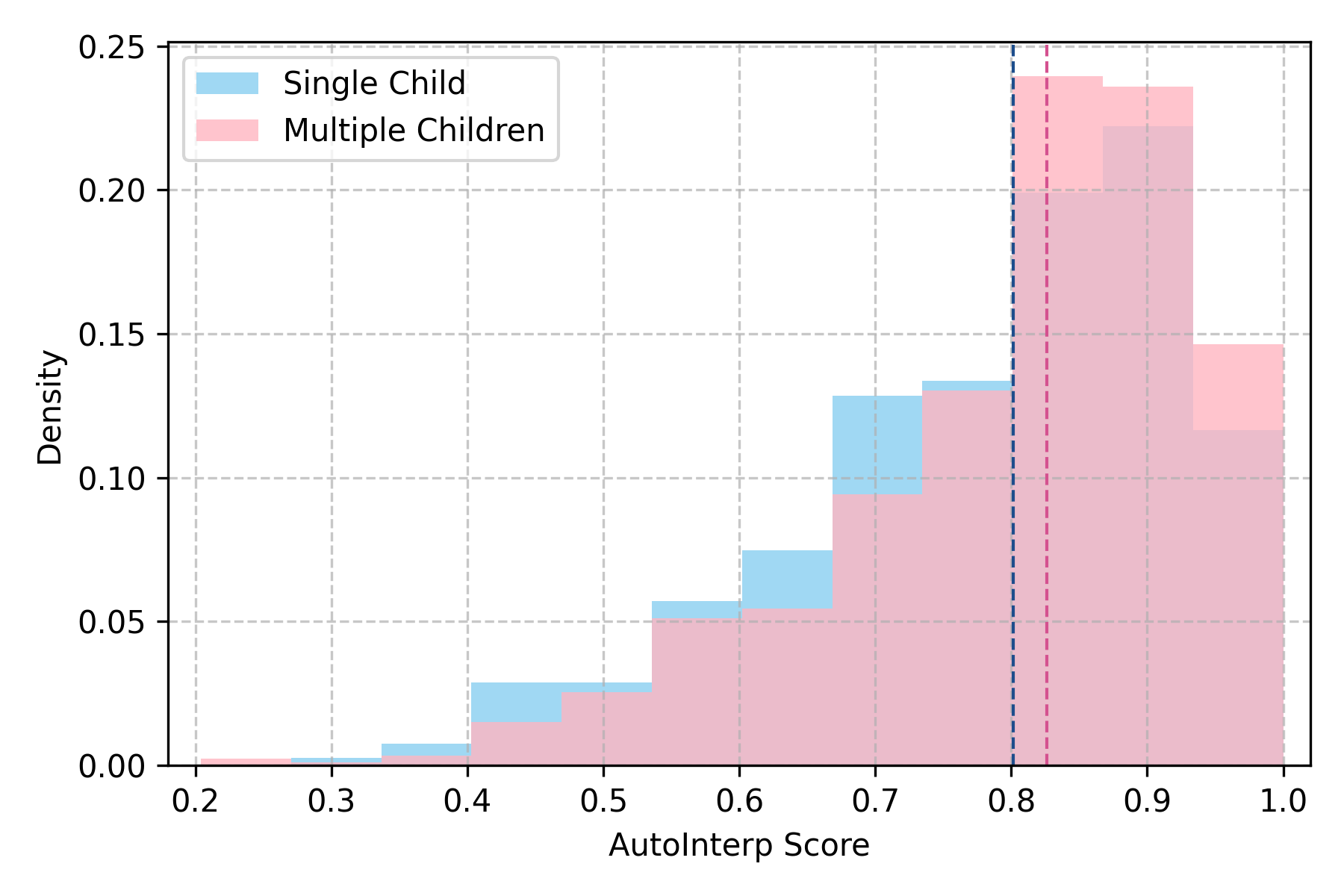}
    \caption{
    \textbf{Interpretability versus structural branching.} Histogram of AutoInterp score of root features with single/multiple children. Feature with multiple children tends to be more interpretable comparing with those with single child. The average AutoInterp score gap is 2.49\%, comparable to the performance gain achieved by quadrupling the dictionary size (from 16k to 64k) in standard SAEs.
    }
    \label{fig:autointerp_gap}
\end{figure}

As mentioned in the main text, \Cref{fig:autointerp_gap} provides a detailed distribution of AutoInterp scores categorized by the branching factor of root features.

\end{document}